\documentclass[10pt,journal,compsoc]{IEEEtran}
\usepackage{graphicx}
\usepackage{epsfig}
\usepackage{amsmath}
\usepackage{amssymb}
\usepackage{tabularx}
\usepackage{multirow}
\usepackage{amssymb}
\usepackage{amsmath}
\usepackage{pifont}
\usepackage{subfigure}
\usepackage{hyperref}
\usepackage[super]{nth}
\newcommand{\cmark}{\ding{51}}%
\newcommand{\xmark}{\ding{55}}%
\usepackage{color}
\usepackage{balance}
\usepackage{booktabs}
\usepackage{overpic}

\newcommand{\etal}{\emph{et al. }}
\newcommand{\eg}{\emph{e.g., }}
\newcommand{\ie}{\emph{i.e., }}
\newcommand\mypara[1]{\vspace{1mm}\noindent\textbf{#1}}

\ifCLASSOPTIONcompsoc
  \usepackage[nocompress]{cite}
\else
  \usepackage{cite}
\fi

\ifCLASSINFOpdf
\else
\fi

\begin{document}
%
\title{Learning by Distillation: A Self-Supervised Learning Framework for Optical Flow Estimation}
%
%
%

\author{Pengpeng Liu,
        Michael R. Lyu, \textit{Fellow, IEEE},
        Irwin King, \textit{Fellow, IEEE},
        and Jia Xu
\IEEEcompsocitemizethanks{\IEEEcompsocthanksitem P. Liu, M. R. Lyu and I. King are with the Department
of Computer Science and Engineering, The Chinese University of Hong Kong, Sha Tin, Hong Kong. P. Liu did mainly the work during an internship at Huya AI. \protect\\
E-mails: \{ppliu, lyu, king\}@cse.cuhk.edu.hk
\IEEEcompsocthanksitem J. Xu is with Huya AI, Guangzhou, China.
E-mail: xujia@huya.com
}
}

%
%


\IEEEtitleabstractindextext{%
\begin{abstract}
We present DistillFlow, a knowledge distillation approach to learning optical flow. DistillFlow trains multiple teacher models and a student model, where challenging transformations are applied to the input of the student model to generate hallucinated occlusions as well as less confident predictions. Then, a self-supervised learning framework is constructed: confident predictions from teacher models are served as annotations to guide the student model to learn optical flow for those less confident predictions. The self-supervised learning framework enables us to effectively learn optical flow from unlabeled data, not only for non-occluded pixels, but also for occluded pixels. DistillFlow achieves state-of-the-art unsupervised learning performance on both KITTI and Sintel datasets.

Our self-supervised pre-trained model also provides an excellent initialization for supervised fine-tuning, suggesting an alternate training paradigm in contrast to current supervised learning methods that highly rely on pre-training on synthetic data. At the time of writing, our fine-tuned models ranked 1st among all monocular methods on the KITTI 2015 benchmark, and outperform all published methods on the Sintel Final benchmark.
More importantly, we demonstrate the generalization capability of DistillFlow in three aspects: framework generalization, correspondence generalization and cross-dataset generalization.

\end{abstract}

}

\maketitle

\IEEEdisplaynontitleabstractindextext

%
\IEEEpeerreviewmaketitle

\IEEEraisesectionheading{\section{Introduction}}
Optical flow, which describes the dense pixel correspondence between two adjacent images, has a wide range of applications such as autonomous driving~\cite{menze2015object}, object tracking~\cite{chauhan2013moving} and video-related tasks~\cite{simonyan2014two,Bonneelsiggraph2015,chen2017coherent,sajjadi2018frame}. Traditional variational approaches~\cite{horn1981determining,brox2004high,brox2011large} formulate optical flow estimation as an energy minimization problem, which are usually computationally expensive~\cite{XRK2017} and not applicable for real-time applications.

Benefited from the development of deep learning, convolutional neural networks (CNNs) have been successfully applied to optical flow estimation~\cite{ilg2017flownet,sun2018pwc,hui18liteflownet}, which achieve comparable or even better performance compared with traditional methods, while running at real-time. Similar to other deep learning tasks, a large amount of labeled training data is required to train these fully supervised CNNs with high performance. However, it is extremely difficult to obtain the ground truth of optical flow for real-world image pairs, especially when there are occlusions. Due to lacking large-scale real-world annotations, existing methods highly rely on pre-training on synthetic labeled datasets~\cite{dosovitskiy2015flownet,mayer2016large}. However, there usually exists a large domain gap between the distribution of synthetic data and natural scenes. In order to train a flow model stably,  we have to carefully follow specific training schedules across different datasets~\cite{ilg2017flownet,hui18liteflownet,sun2018pwc,Hur:2019:IRR,barhaim2020scopeflow,zhao2020maskflownet}. This raises our first question: \textit{can we simplify the training procedure and remove the need of synthetic dataset for supervised optical flow estimation?}

\begin{figure}[t]
  \centering
  \includegraphics[width=0.42\textwidth]{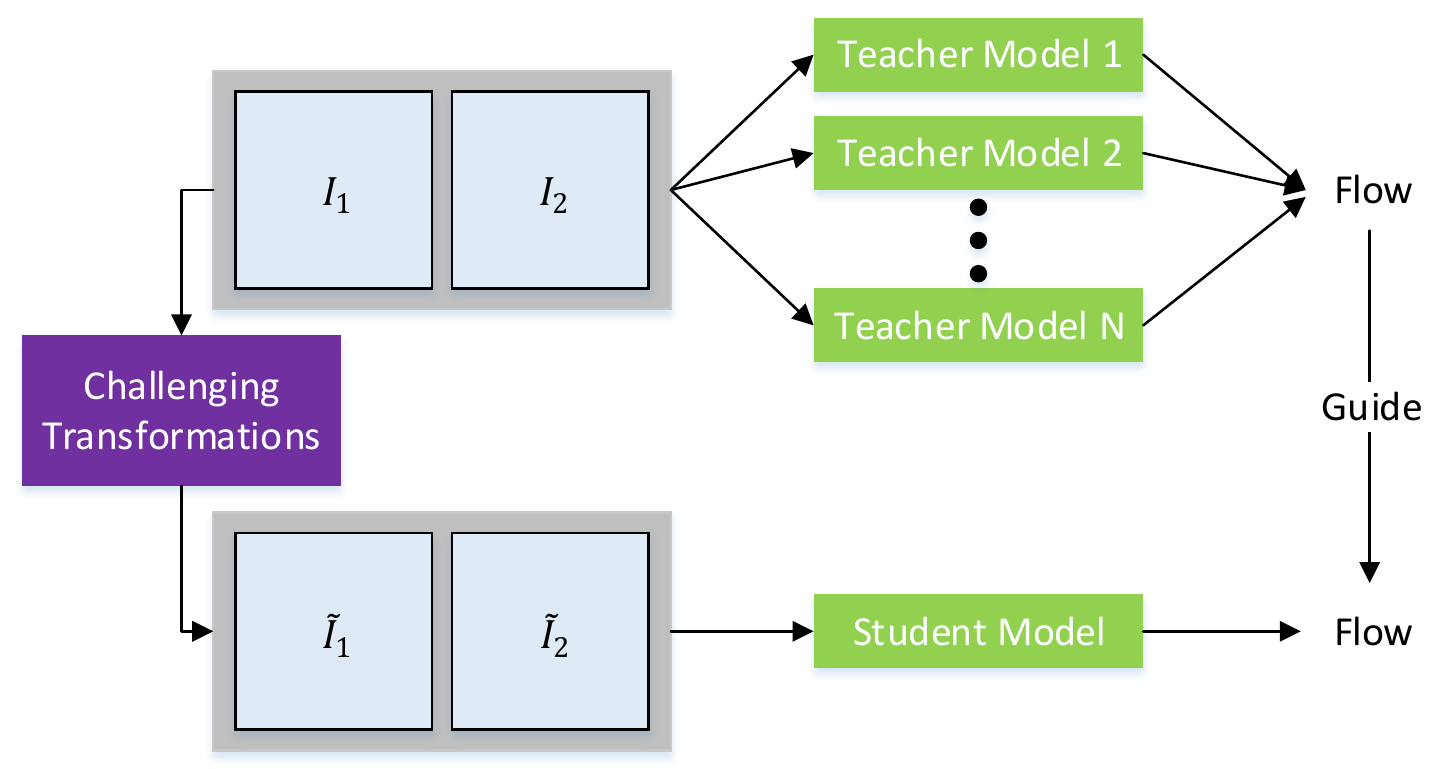}
  \caption{Framework illustration. We distill confident optical flow estimations from teacher models (stage 1) to guide the flow learning of the student model (stage 2) under different challenging transformations.}
  \label{figure:teaser}
  \vspace{-1ex}
\end{figure}

\begin{figure*}[t]
  \centering
  \includegraphics[width=0.94\textwidth]{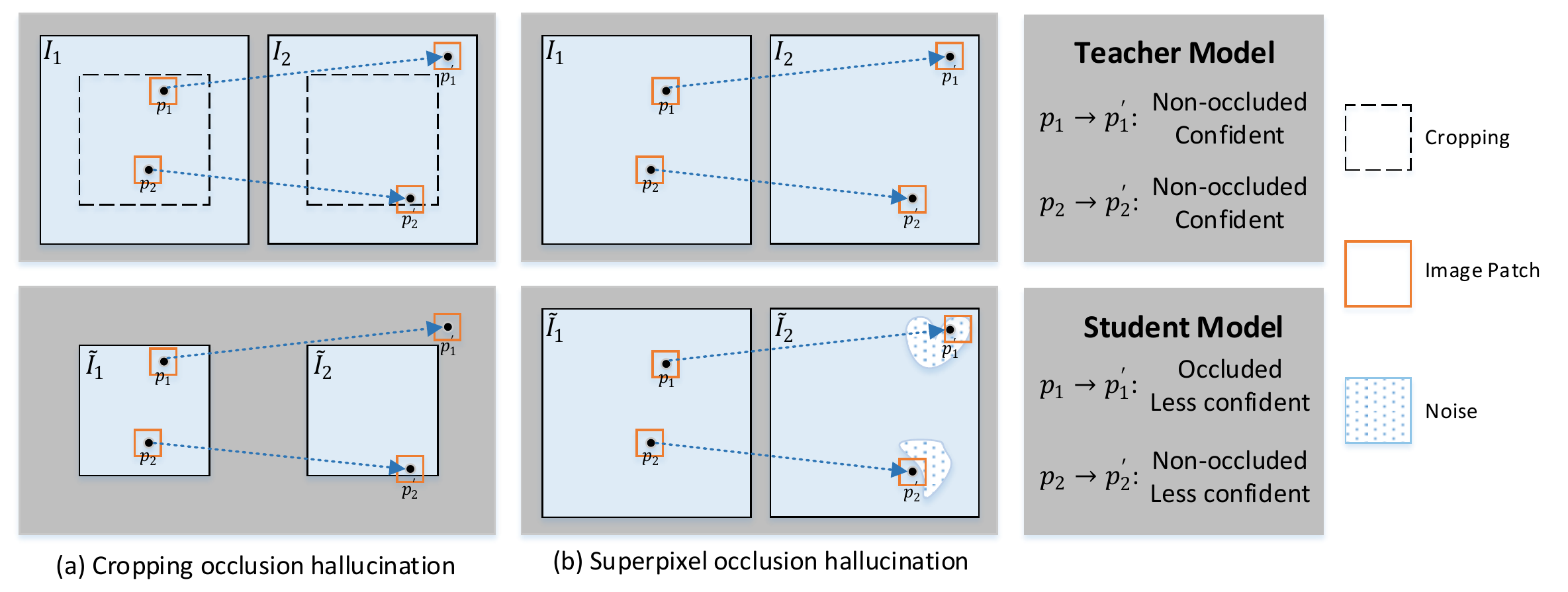}
  \vspace{-2ex}
  \caption{Occlusion hallucination scheme. The scheme creates hand-crafted occlusions, \eg pixel $p_1$ is non-occluded from $I_1$ to $I_2$ but becomes occluded from $\widetilde{I}_1$ to $\widetilde{I}_2$ ($p_1^{'}$ moves out of the image boundary for case (a) and is covered by noise for case (b)). The scheme also creates less confident predictions, \eg though pixel $p_2$ is non-occluded, its patch similarity between $\widetilde{I}_1$ and $\widetilde{I}_2$ is smaller than patch similarity between $I_1$ and $I_2$ due to the partially missing regions.}
  \label{figure:distillation}
  \vspace{-1ex}
\end{figure*}

Another promising direction is to develop unsupervised methods~\cite{jason2016back,ren2017unsupervised,Meister:2018:UUL,wang2018occlusion,Janai2018ECCV}, where unlimited image sequences are readily available. The basic idea is based on image warping, where the target image is warped toward the reference image according to the estimated flow, then a photometric loss is defined to minimize the difference between the reference image and the warped target image. However, brightness constancy assumption does not hold for occluded regions. Therefore, such photometric loss only works well for non-occluded pixels, but turns to provide misleading information for occluded pixels. To alleviate the above issue, recent methods propose to exclude occluded pixels when computing the photometric loss or employ additional spatial and temporal smoothness terms to regularize the flow estimations~\cite{Meister:2018:UUL,wang2018occlusion,Janai2018ECCV}. However, all these methods rely on hand-crafted energy terms to regularize flow estimation,  lacking the key capability to effectively learn optical flow of occluded pixels.  As a result, there is still a large performance gap comparing these methods with state-of-the-art fully supervised methods. This raises our second question: \textit{can we effectively learn optical flow with occlusions without relying on any labeled data and reduce the performance gap?}

In this paper, we address the above two questions with a two-stage self-supervised learning approach based on knowledge distillation. As shown in Figure~\ref{figure:teaser}, the proposed framework contains two kinds of models: teacher model and student model, and two kinds of distillation: model distillation (stage 1) and data distillation (stage 2). In stage 1, we train teacher models following a similar training protocol as previous methods~\cite{Meister:2018:UUL,wang2018occlusion,Janai2018ECCV} with occlusion handling, aiming at estimating accurate optical flow for those non-occluded pixels. We employ model distillation to ensemble predictions from multiple teacher models, which can reduce the variance of single teacher model prediction. The confident predictions from teacher models serve as annotations to guide the learning of the student model in the second stage. For data distillation, we create challenging transformations to the input image pairs, which are used to generate hallucinated occlusions as well as less confident predictions for self-supervision. As shown in Figure~\ref{figure:distillation}, we show occlusion hallucination techniques that generate hand-crafted occlusions, \eg pixel $p_1$ is non-occluded from $I_1$ to $I_2$ but becomes occluded from $\widetilde{I}_1$ to $\widetilde{I}_2$. Then we can let teacher models guide the student model to learn the optical flow of these hand-crafted occluded pixels. As a result, the distillation scheme enables our student model to effectively learn the optical flow of occluded pixels. The occlusion hallucination techniques also create less confident predictions, \eg although $p_2$ is non-occluded, the patch similarity of pixel $p_2$ between $\widetilde{I}_1$ and $\widetilde{I}_2$ is smaller than the patch similarity between $I_1$ and $I_2$, due to the partially missing regions. In this case, the distillation scheme helps our student model learn optical flow of non-occluded pixels more effectively. Apart from these occlusion hallucination techniques, we create other challenging transformations, including geometric transformations (\eg scaling) and color transformations (\eg overexposure and underexposure). Overall, our distillation scheme lets confident predictions from teacher models to supervise less confident predictions from the student model, enabling the student model to have the improved ability to learn optical flow of both occluded and non-occluded pixels. Our method, termed as DistillFlow, outperforms all previous unsupervised methods by a large margin on all datasets, and is comparable with fully supervised methods, which significantly reduces the performance gap. Hence, the problem in question 2 is successfully addressed.

More importantly, our self-supervised pre-trained model provides an excellent initialization for supervised fine-tuning. We can use self-supervised pre-training on unlabeled data to replace pre-training on multiple synthetic datasets. We also extend the distillation idea from unsupervised learning to semi-supervised learning, which further improves the fine-tuning performance. At the time of writing, our fine-tuned models outperform all monocular methods on the KITTI 2015 benchmark, and outperform all published methods on the Sintel Final benchmark.
Hence, the problem in question 1 is successfully addressed.

Furthermore, we demonstrate the generalization capability of DistillFlow in three aspects: framework generalization, correspondence generalization and cross-dataset generalization. In framework generalization, we first show that our knowledge distillation framework is applicable to different network structures (\eg PWC-Net~\cite{sun2018pwc}, FlowNetS~\cite{dosovitskiy2015flownet} and FlowNetC~\cite{dosovitskiy2015flownet}), then extend the knowledge distillation idea to semi-supervised learning. For correspondence generalization, we directly use our self-supervised flow model trained on monocular videos to estimate stereo disparity. Surprisingly, DistillFlow achieves comparable performance with state-of-the-art unsupervised stereo matching methods on KITTI datasets. For cross-data generalization, we evaluate the performance of the model trained on another dataset (\eg Sintel~$\to$~KITTI and KITTI~$\to$~Sintel), and show that DistillFlow still achieves comparable performance with previous methods.

We summarize our main contributions as follows:
\begin{itemize}
\item We present a two-stage self-supervised learning approach based on knowledge distillation to effectively learn optical flow of both occluded and non-occluded pixels from unlabeled data. Our method significantly outperforms previous unsupervised methods, especially for occluded pixels. 

\item We improve the training protocol compared with our previous works DDFlow~\cite{Liu:2019:DDFlow}, SelFlow~\cite{Liu:2019:SelFlow} and Flow2Stereo~\cite{Liu:2020:Flow2Stereo} by adding spatial regularizer and model distillation.

\item Our self-supervised pre-trained models provide excellent initializations for supervised fine-tuning, which removes the reliance of pre-training on multiple synthetic datasets. After fine-tuning, we achieve state-of-the-art supervised learning performance.

\item  We demonstrate the generalization capability of DistillFlow in three aspects: framework generalization, correspondence generalization and cross-dataset generalization.

\end{itemize}

\section{Related Work}
\mypara{Traditional Flow Methods.}
Traditional variational methods formulate optical flow estimation as an energy minimization problem based on brightness constancy and spatial smoothness~\cite{horn1981determining,sun2010secrets}. These methods work well for small motion, but usually fail when displacements are large. Later works~\cite{brox2011large,weinzaepfel2013deepflow} integrate feature matching  to tackle this issue. Specifically, they first find sparse feature correspondences to initialize flow estimation and then refine it in a coarse-to-fine manner. The seminal work EpicFlow \cite{revaud2015epicflow} interpolates dense flow from sparse matches and has become a widely used post-processing pipeline. There are also some works that use temporal information over multiple frames to improve the robustness and accuracy by adding temporal constraints, such as constant velocity~\cite{janai2017slow,kennedy2015optical,sun2010layered}, constant acceleration~\cite{volz2011modeling,black1991robust} and so on. Recently, \cite{bailer2017cnn,XRK2017} use convolutional neural networks (CNNs)  to learn a feature embedding for better matching and have demonstrated superior performance. However, these methods are often computationally expensive and can not be trained end-to-end. In this paper, we use CNNs to directly estimate optical flow in an end-to-end manner, which is very efficient.

\mypara{Supervised Flow Methods.}
Inspired by the development of deep neural networks, CNNs have been successfully applied to optical flow estimation. The pioneering work FlowNet \cite{dosovitskiy2015flownet} proposes two types of CNN, FlowNetS and FlowNetC, which take two consecutive images as input and output a dense optical flow map. The follow-up FlowNet 2.0~\cite{ilg2017flownet} stacks several basic FlowNet models and refines the flow iteratively, which significantly improves accuracy. SpyNet~\cite{ranjan2017optical} proposes a light-weight network architecture by employing image warping at different scales in a coarse-to-fine manner. However, its performance is behind the state-of-the-art. PWC-Net~\cite{sun2018pwc} and LiteFlowNet~\cite{hui18liteflownet} propose to warp CNN features instead of images at different scales and introduce cost volume construction, achieving state-of-the-art performance with compact model size. They were further improved by only using a single network block with shared weights to iteratively refine flow at different scales and adding occlusion reasoning~\cite{Hur:2019:IRR}. VCN~\cite{yang2019volumetric} introduces efficient volumetric networks for dense 2D correspondence matching by exploring high-dimensional invariance during cost volume computation. MaskFlowNet~\cite{zhao2020maskflownet} proposes an asymmetric occlusion-aware feature matching module, which masks out those occluded regions after feature warping. ScopeFlow~\cite{barhaim2020scopeflow} introduces an improved training protocol by fully utilizing cropping randomly sized scene scopes. However, due to lacking of real-world ground truth optical flow, all above supervised learning methods highly rely on pre-training on synthetic datasets (\eg FlyingChiars~\cite{dosovitskiy2015flownet} and FlyingThings3D~\cite{mayer2016large}) and follow specific training schedules. In this paper, we propose to employ self-supervised pre-training on unlabeled image sequences to achieve excellent initializations, which remove the reliance of pre-training on synthetic datasets.

\begin{figure*}[t]
  \centering
  \includegraphics[width=1\textwidth]{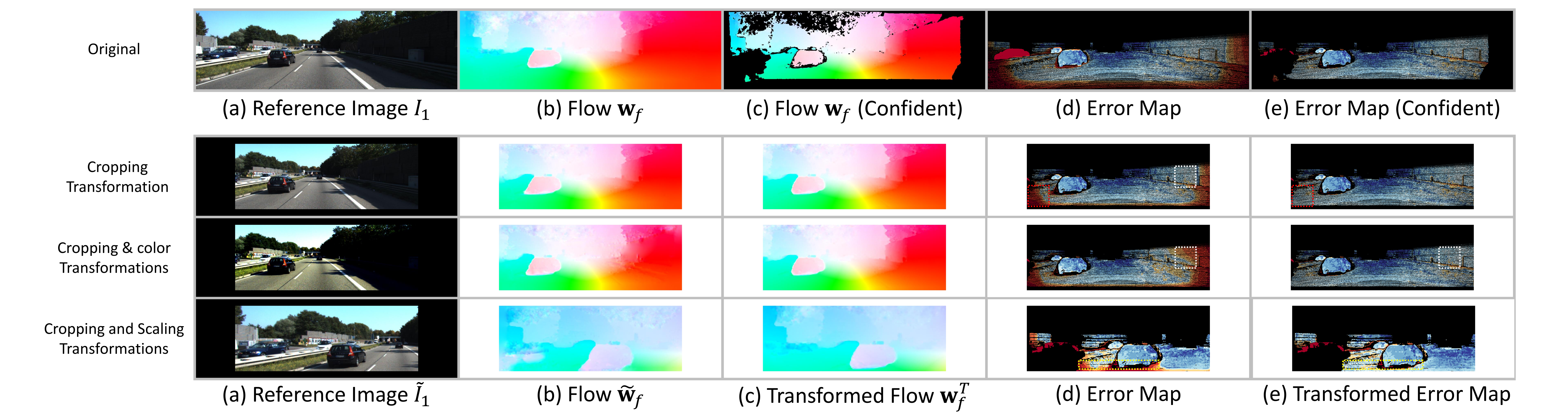}
  \vspace{-4ex}
  \caption{Knowledge distillation examples. The redder the color in the error map, the greater the error. In the (d) of the first row, flow $\textbf{w}_f$ has many erroneous pixels; however the confident flow predictions after forward-backward consistency check in (c) are mostly reliable (as shown in (e)). After creating challenging transformations to the input (\eg row 2-4), the flow predictions by the student model are usually less confident than the transformed predictions $\textbf{w}_f^T$ from confident flow, \eg rectangle regions in the error maps. We only use confident predictions in (c) of row 1 to guide the learning of the student model. In general, $\textbf{w}_f^T$ shall be sparse as in (c) of row 1. For better visual comparison with $\widetilde{w}_f$, we show transformed results from (b) of row 1.}
  \label{figure:distillation_example}
  \vspace{-1ex}
\end{figure*}

\mypara{Unsupervised Flow Methods.}
Due to lacking ground truth optical flow for natural image sequences, recent studies turn to formulate optical flow estimation as an unsupervised learning problem based on the brightness constancy and spatial smoothness assumption~\cite{jason2016back,ren2017unsupervised}. The basic idea is based on image warping to achieve view synthesize, and then define a photometric loss measure the difference between the reference image and the synthesized warped target image. However, brightness constancy does not hold for occluded pixels, therefore \cite{Meister:2018:UUL,wang2018occlusion,Janai2018ECCV} propose to detect occlusion and exclude occluded pixels when computing photometric loss. EpipolarFlow~\cite{zhong2019epiflow} proposes to incorporate global geometric epipolar constraints into  network learning to improve performance. There are also works that propose to jointly learn flow and depth from monocular videos~\cite{zhou2017unsupervised,zou2018df,yin2018geonet,ranjan2019competitive,liu2019unsupervised} or jointly learn flow and disparity from stereoscopic videos~\cite{lai2019bridging,wang2019unos}. Despite promising progress, they still lack the key ability to effectively learn optical flow of occluded pixels and their performance is far behind state-of-the-art supervised learning methods. In this paper, we propose a knowledge distillation based self-supervised learning method to learn optical flow of both occluded and non-occluded pixels in an unsupervised manner. 

\mypara{Unsupervised Stereo Methods.}
Optical flow estimation and stereo matching can be viewed as a unified problem: correspondence matching. For rectified stereo image pairs, the stereo disparity can be regarded as a special case of optical flow in the horizontal direction. Therefore, our method is also related to a large body of unsupervised stereo learning methods, including image synthesis and warping with depth estimation~\cite{garg2016unsupervised}, left-right consistency~\cite{Godard_2017_CVPR,Zhou_2017_ICCV,Guo_2018_ECCV}, employing  additional semantic information~\cite{yang2018segstereo}, cooperative learning~\cite{li2018occlusion},  self-adaptive fine-tuning~\cite{tonioni2017unsupervised,zhong2018open,tonioni2019real}. Different from all these methods that train stereo models on stereo image pairs, we directly use our optical flow model trained on monocular videos to estimate disparity (only keep the horizontal component of flow). We demonstrate that our self-supervised flow model has the strong generalization capability to find correspondences.

\mypara{Knowledge Distillation and Self-Supervised Learning.}
Our work is closely related to the family of self-supervised learning methods, where the supervision signal is purely generated from the data itself.
It is widely used for learning feature representations from unlabeled data~\cite{jing2020self}. A pretext task is usually employed, such as image inpainting~\cite{pathak2016context}, image colorization~\cite{larsson2017colorization}, solving Jigsaw puzzles~\cite{noroozi2016unsupervised}. Pathak~\etal~\cite{pathak2017learning} propose to explore low-level motion-based cues to learn feature representations without manual supervision. Doersch~\etal~\cite{doersch2017multi} combine multiple self-supervised learning tasks to train a single visual representation. To obtain more reliable supervision signals, we employ model distillation~\cite{hinton2015distilling} to ensemble predictions of several teacher models. Then we make use of the domain knowledge of optical flow and employ data distillation for self-supervision. Our data distillation is different from \cite{radosavovic2018data}, which ensembles predictions from a single model applied to multiple transformations of unlabeled image pairs as annotations. Instead, we create challenging image transformations to create hallucinated occlusions and less confident predictions to enable effective self-supervision.

\section{Method}
In this section, we illustrate our self-supervised learning framework based on knowledge distillation. We train two types of CNNs (multiple teacher models and a student model) with the same network architecture. Only the student model is needed during testing. The self-supervised learning framework contains two stages: in stage 1, we train teacher models to obtain confident flow predictions; in stage 2, we distill confident predictions from the teacher models to guide the student model to learn optical flow of both occluded and non-occluded pixels. We introduce two variants in stage 2, where variant 1 is from the occlusion view and variant 2 is from the confidence view. In principle, DistillFlow can use any backbone network to learn optical flow. In our implementation, we adopt IRR-PWC~\cite{Hur:2019:IRR}, which is a variant of PWC-Net~\cite{sun2018pwc} with weight sharing across different levels. We additionally add dilated convolution to increase the receptive field. Before describing our method in detail, we first define our notations.

\subsection{Notation}
For our teacher models, we denote $I_1$, $I_2$ $\in\mathbb{R}^{H \times W \times 3}$ for two consecutive RGB images, where $H$ and $W$ are height and width respectively. Our goal is to estimate the forward optical flow $\textbf{w}_{f} \in \mathbb{R}^{H \times W \times 2} $ from $I_1$ to $I_2$. After obtaining  $\textbf{w}_{f}$, we can warp $I_2$ towards $I_1$ to get the warped image $I_{2 \to 1}^w$. Here, we also estimate the backward optical flow $\textbf{w}_b$ from $I_2$ to $I_1$ and a backward warped image $I_{1 \to 2}^w$. Since there are many cases where one pixel is only visible in one image but not visible in the other image, namely occlusion, we denote $O_f$, $O_b$ $\in\mathbb{R}^{H \times W \times 1}$ as the forward and backward occlusion map respectively. The occlusion map is a binary mask, where value 1 denotes that the pixel in that location is occluded and value 0 denotes non-occluded.

After creating challenging transformations to the input image pairs, the input images to the student model are denoted as $\widetilde{I}_1$, $\widetilde{I}_2$ $\in\mathbb{R}^{h \times w \times 3}$. Similarly, optical flow, occlusion map and confidence map from teacher models need to perform corresponding transformations as input images. We use $\textbf{w}_f^T$, $\textbf{w}_b^T$, $O_f^T$, $O_b^T$, $M_f^T$ and $M_b^T$ to denote their transformed results.

Our student model follows similar notations. The student network takes $\widetilde{I}_1$, $\widetilde{I}_2$ as input, and produces optical flow $\widetilde{\textbf{w}}_{f}$, $\widetilde{\textbf{w}}_{b}$, warped images $\widetilde{I}_{2 \to 1}^w$, $\widetilde{I}_{1 \to 2}^w$, occlusion maps $\widetilde{O}_{f}$, $\widetilde{O}_{b}$.

For stage 2 variant 1, knowledge distillation is performed from occlusion view. Therefore, we define another occlusion map $O_f^{'}$ and $O_b^{'}$ to denote hallucinated occlusions (\ie hand-crafted occlusions). Hallucinate occlusion map is computed from the transformed occlusion mask $O^T$ (in our teacher models) and occlusion mask $\widetilde{O}$ (in our student model).

For stage 2 variant 2, knowledge distillation is performed from the confidence view. Therefore, we define confidence maps $M_f^T$ and $M_b^T$ to denote which pixels can be accurately estimated by our teacher models after transformations. For confidence maps, value 1 denotes that flow prediction of that pixel is confident and value 0 denotes not confident.

\subsection{Stage 1: Unsupervised Flow Learning}
To train our teacher models in an unsupervised manner, we swap the image pairs in our input to produce both forward flow $\textbf{w}_f$ and backward flow $\textbf{w}_b$. After that, we estimate an occlusion map based on the forward-backward consistency prior~\cite{sundaram2010dense,Meister:2018:UUL}. That is, for non-occluded pixels, the forward flow should be the inverse of the backward flow at the corresponding pixel in the second image. We consider pixels as occluded when the mismatch between forward flow and backward flow is too large. Take forward occlusion map as an example, we first compute the reversed forward flow as follow:
\begin{equation}
  \hat{\textbf{w}}_f(\textbf{p}) = \textbf{w}_b({\textbf{p}+\text{w}_f}(\textbf{p})),
\end{equation}
where $\textbf{p}$ is a pixel in the first image $I_1$. A pixel is considered occluded (\ie $O_f(\textbf{p})=1$) when the following constraint is violated:
\begin{equation}
|\textbf{w}_f(\textbf{p}) + \hat{\textbf{w}}_f(\textbf{p})|^2 < \alpha_1 (|\textbf{w}_f(\textbf{p})|^2+|\hat{\textbf{w}}_f(\textbf{p})|^2) + \alpha_2,
\end{equation}
where we set $\alpha_1$ = 0.01, $\alpha_2$ = 0.5 for all our experiments. Backward occlusion map $\textbf{w}_b$ is computed in the same way.

Unsupervised flow estimation is based on brightness constancy and spatial smoothness assumption. We use photometric loss $L_{pho}$ and edge-aware smoothness loss $L_{smo}$ for the above two assumptions. Photometric loss measures the difference between the reference image and the warped target image. Take forward flow $\textbf{w}_f$ as an example, we can use $\textbf{w}_f$ to warp $I_2$ to reconstruct $I_1$:
\begin{equation}
I_{2 \to 1}^w(\textbf{p}) = I_2(\textbf{p} + \textbf{w}_f(\textbf{p})).
\end{equation}
Photometric loss $L_{pho}$ only makes sense for non-occluded pixels, which is defined as follows:
\begin{multline}
  L_{pho} = \sum{\psi(I_1-I_{2 \to 1}^{w}) \odot (1 - O_f)} / \sum{(1 - O_f)} \\
  + \sum{\psi(I_2-I_{1 \to 2}^{w}) \odot (1-O_b)} / \sum(1 - O_b)
    \label{eq:photometric_loss}
\end{multline}
where $\psi(x) = (|x|+\epsilon)^q$ is a robust loss function, $\odot$ denotes the element-wise multiplication. During our experiments, we set $\epsilon=0.01$, $q=0.4$.

Photometric loss is not informative in homogeneous regions, therefore existing unsupervised methods usually add a smoothness loss to regularize the flow~\cite{Meister:2018:UUL,wang2018occlusion}. Smoothness can be regarded as a regularizer for occluded pixels, since it makes the prediction of occluded pixels similar to the neighborhood pixels. Here we adopt an edge-aware smoothness loss weighted by the image gradient:
\begin{multline}
  L_{smo} = \frac {1} {H \times W} \sum_{\textbf{p}} |e^{- \beta \nabla{I_1}(\textbf{p})}|^T \cdot |\nabla \textbf{w}_f(\textbf{p})| \\
  + \frac {1} {H \times W} \sum_{\textbf{p}} |e^{- \beta \nabla{I_2}(\textbf{p})}|^T \cdot |\nabla \textbf{w}_b(\textbf{p})|,
  \label{eq:smoothness_loss}
\end{multline}
where $\nabla$ denotes gradient, $T$ represents transpose and $\beta$ is a factor to control the smoothness effect on edges. We set $\beta=10$ in our experiment.

The final loss to train teacher models in stage 1 is the combination of $L_{pho}$ and $L_{smo}$:
\begin{equation}
  L_1 = L_{pho} + 0.1 * L_{smo}.
  \label{eq:stage_1}
\end{equation}
\subsection{Stage 2: Self-Supervised Flow Learning with Knowledge Distillation}
Since photometric loss does not make sense for occluded pixels, prior unsupervised methods lack the key ability to effectively learn optical flow of occluded pixels. To tackle this issue, we distill confident predictions from our teacher models, and use them to generate input/output data for our student model by creating challenging transformations. Next, we first introduce our occlusion hallucination techniques, then describe challenging transformations employed in DistillFlow, finally we explain two variants of knowledge distillation.

\subsubsection{Occlusion Hallucination}
Figure~\ref{figure:distillation} demonstrates two kinds of occlusion hallucination techniques used in DistillFlow: random cropping and random superpixel noise injection. In both Figure~\ref{figure:distillation} (a) and (b), suppose pixel $p_1$ is non-occluded from $I_1$ to $I_2$ and pixel $p_1^{'}$ in its corresponding pixel.  After creating challenging transformation to $I_1$ and $I_2$, $p_1$ becomes occluded from $\widetilde{I}_1$ to $\widetilde{I}_2$, because $p_1^{'}$ moves out of image boundary in (a) and is covered by noise in (b). We call the above procedures of creating hand-crafted occlusions as occlusion hallucination.

Even though $p_1$ becomes occluded from $\widetilde{I}_1$ to $\widetilde{I}_2$, the location of its corresponding pixel $p_1^{'}$ does not change. As a result, the flow of $p_1$ does not change during occlusion hallucination. This is the basic assumption of our knowledge distillation idea. Since $p_1$ is non-occluded from $I_1$ to $I_2$, our teacher models can accurately estimate its flow; however, $p_1$ becomes occluded from $\widetilde{I}_1$ to $\widetilde{I}_2$, therefore our student model cannot accurately estimate its flow anymore. Luckily, we can distill confident flow estimation of $p_1$ from teacher models to guide the student model to learn the flow of $p_1$ from $\widetilde{I}_1$ to $\widetilde{I}_2$. This explains why our knowledge distillation approach enables our student model to effectively learn optical flow of occluded pixels.

Strategy in (a) generates hallucinated occlusions near the image boundary. However, for occlusion elsewhere (\eg motion boundary of objects), it is not so effective. Strategy in (b) can generate hallucinated occlusions in a wider range of cases. This is because the shape of a superpixel is usually random and superpixel edges are often part of object boundaries, which is consistent with the real-world cases. We can choose several superpixels at different locations to cover more occlusion cases. The combination of (a) and (b) can generate a variety of hallucinated occlusions.

\subsubsection{Challenging Transformations} \label{section:challenging_transformations}
In this part, we show that knowledge distillation also helps learn the optical flow of non-occluded pixels. We introduce three kinds of challenging transformations: occlusion hallucination-based transformations, geometric transformations and color transformations.

\textbf{Occlusion-hallucination based transformations.}
When searching pixel correspondences, we care about not only the color of specific pixels, but also the color of their neighbors or context, that is, image patch similarity. In Figure~\ref{figure:distillation}, $p_2$ is non-occluded both from $I_1$ to $I_2$ and from $\widetilde{I}_1$ to $\widetilde{I}_2$, and $p_2^{'}$ is its corresponding pixel. When considering the image patches around pixel $p_2$ and $p_2^{'}$, patch similarity from $\widetilde{I}_1$ to $\widetilde{I}_2$ is obviously smaller than patch similarity from ${I}_1$ to ${I}_2$, due to the partially missing regions (part regions move out of the image boundary in (a) and part regions are covered by noise in (b)). In this case, the flow estimation of $p_2$ by teacher models is more confident than the student model. Then, we can use the confident predictions from teacher models to guide the student model to learn the flow of $p_2$. This explains why knowledge distillation also improves the optical flow of non-occluded pixels with occlusion hallucination based transformations.

\textbf{Geometric transformations.}
Geometric transformations include cropping, scaling, rotation, translation and so on, which are defined by 6 parameters as in the affine transformation from Spatial Transformer Network~\cite{jaderberg2015spatial}. Cropping used in occlusion hallucination is just one kind of geometric transformation. Actually, other kinds of geometric transformations are also effective as long as they can create challenging scenes, where flow predictions become less confident after transformations. Take scaling as an example, suppose we downsample $I_1$ and $I_2$ as input to the student model. In general, image information will lose during the downsampling operation, which makes the student model difficult to predict flow. Therefore, the flow prediction by the student model is less confident than teacher models.

\textbf{Color transformations.}
Color transformations represent those transformations that change the appearance of images, such as changing contrast, brightness, saturation, hue, etc. Though such transformations do not create hallucinated occlusions, they can create challenging scenes. For example, generating images with overexposure and underexposure changes the image appearance, and decreasing image contrast makes pixels less distinguishable.

Overall, the purpose of creating challenging transformations is to create hallucinated occlusion or less confident predictions, so that knowledge distillation can be effectively employed. Figure~\ref{figure:distillation_example} shows a real-world example of different transformations. The raw flow predictions from teacher models have many erroneous pixels, but confident predictions after forward-backward consistency check are mostly correct ((d) $\to$ (e) in row 1). After creating challenging transformations, many confident predictions become less confident (\eg rectangle regions in (d) and (e)).

\subsubsection{Knowledge Distillation}
Knowledge distillation is performed at stage 2 to train our student model in the self-supervision framework. We first introduce two variants of knowledge distillation, then make comparisons between them.

\textbf{Stage 2 variant 1: from the occlusion view.} As illustrated in the occlusion hallucination section, there exist new occlusions after creating challenging transformations. Hallucinated occlusion maps $O_f^{'}$ and $O_b^{'}$ are computed as follows:
\begin{equation}
\left\{
             \begin{array}{lr}
             O_f^{'} = \text{min}(\text{max}(\widetilde{O}_f - O_f^T, 0), 1) &  \\
             O_b^{'} = \text{min}(\text{max}(\widetilde{O}_b - O_b^T, 0), 1). &
             \end{array}
\right.
\end{equation}
Then the pixels in $\widetilde{I}_1$, $\widetilde{I}_2$ can be divided into three types: old occluded pixels (pixels that are occluded from $I_1$ to $I_2$), hallucinated occlusions (pixels that are non-occluded from $I_1$ to $I_2$ but become occluded from $\widetilde{I}_1$ to $\widetilde{I}_2$), and non-occluded pixels both from $I_1$ to $I_2$ and from $\widetilde{I}_1$ to $\widetilde{I}_2$).

As shown in Figure 2, we define the loss for occluded pixels on hallucinated occlusions:
\begin{multline}
L_{occ}  = \sum{\psi(\textbf{w}_f^T-\widetilde{\textbf{w}}_f) \odot O_f^{'}} / \sum{O_f^{'}} \\
  +  \sum{\psi(\textbf{w}_b^T-\widetilde{\textbf{w}}_b) \odot O_b^{'}} / \sum{O_b^{'}},
\end{multline}
where $\textbf{w}_f^T$ and $\textbf{w}_b^T$ are the transformed flow of $\textbf{w}_f$ and $\textbf{w}_b$ from teacher models.

Photometric loss for non-occluded pixels and edge-aware smoothness loss are computed in the same way as the teacher models. The final loss to train our student model in stage 2 variant 1 is the combination of $L_{pho}$, $L_{occ}$ and $L_{smo}$:
\begin{equation}
  L_{2 \_ 1} = L_{pho} + L_{occ} + 0.1 * L_{smo}.
  \label{eq:stage_2_1}
\end{equation}
\textbf{Stage 2 variant 2: from the confidence view.}
In stage 1, forward-backward consistency check is employed to detect whether the pixel is occluded or not. However, this brings in errors because many pixels are still non-occluded even when they violate this principle, and vice versa. Instead, it would be more proper to call those pixels confident if they pass the forward-backward consistency check. From this point of view, the key point of knowledge distillation is to let confident predictions to supervise those less confident predictions. We define confidence map $M$ as the reverse of occlusion map $O$:
\begin{equation}
\left\{
             \begin{array}{lr}
             M_f^T = 1 - O_f^T &  \\
             M_b^T = 1 - O_b^T &
             \end{array}
\right.
\end{equation}
When creating challenging transformations, both occluded regions and non-occluded regions become more challenging. During the self-supervised learning stage, the student model is able to handle more challenging conditions. As a result, its performance improves not only for those occluded pixels, but also for non-occluded pixels.

We define knowledge distillation loss $L_{dis}$ as follows:
\begin{multline}
L_{dis}  = \sum{\psi(\textbf{w}_f^T-\widetilde{\textbf{w}}_f) \odot M_f^T} / \sum{M_f^T} \\
  +  \sum{\psi(\textbf{w}_b^T-\widetilde{\textbf{w}}_b) \odot M_b^T} / \sum{M_b^T}.
\end{multline}
The final loss to train our student model in stage 2 variant 2 is the combination of  $L_{dis}$ and $L_{smo}$:
\begin{equation}
  L_{2 \_ 2} = L_{dis} + 0.1 * L_{smo}.
  \label{eq:stage_2_2}
\end{equation}
\textbf{Comparison of two variants.}
\begin{figure*}[t]
 \centering
 \subfigure[Reference Image]{
 \begin{minipage}[t]{0.196\textwidth}
 \includegraphics[width=1\textwidth]{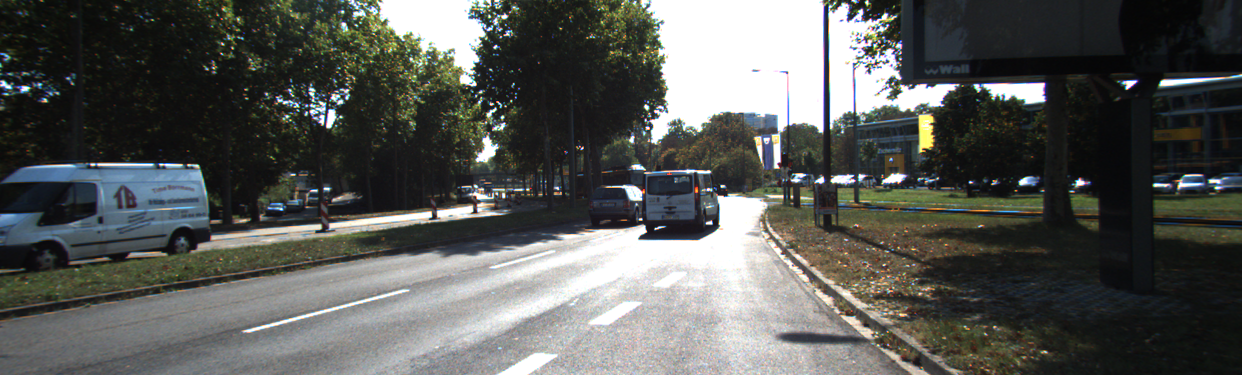} \\
 \includegraphics[width=1\textwidth]{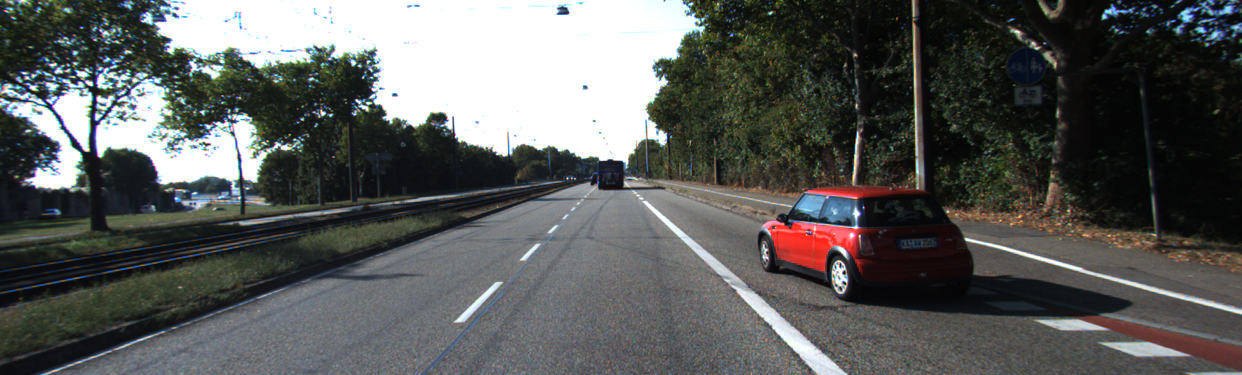} \\
 \includegraphics[width=1\textwidth]{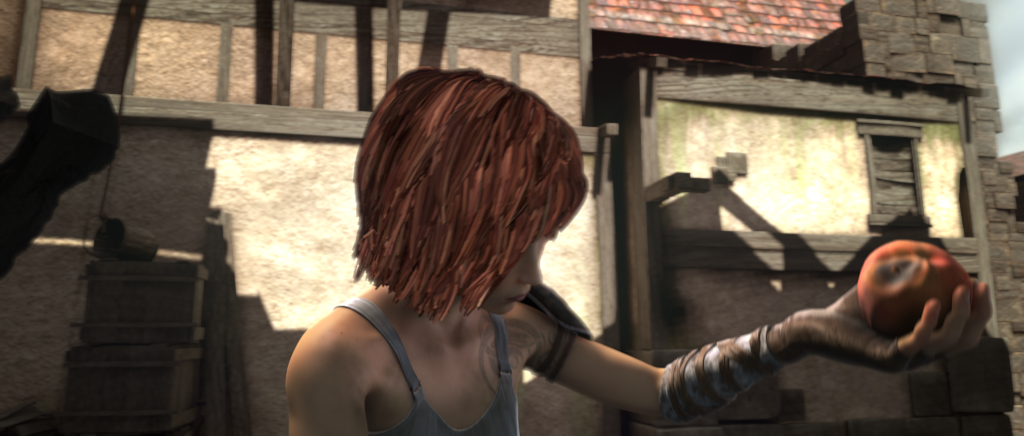} \\
 \includegraphics[width=1\textwidth]{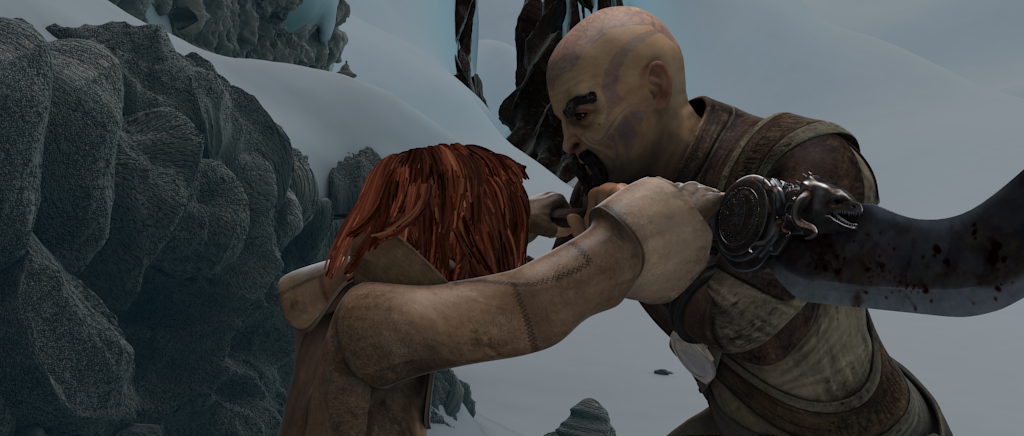}
 \end{minipage}
 }
 \hspace{-0.02\textwidth}
 \subfigure[GT Flow]{
 \begin{minipage}[t]{0.196\textwidth}
 \includegraphics[width=1\textwidth]{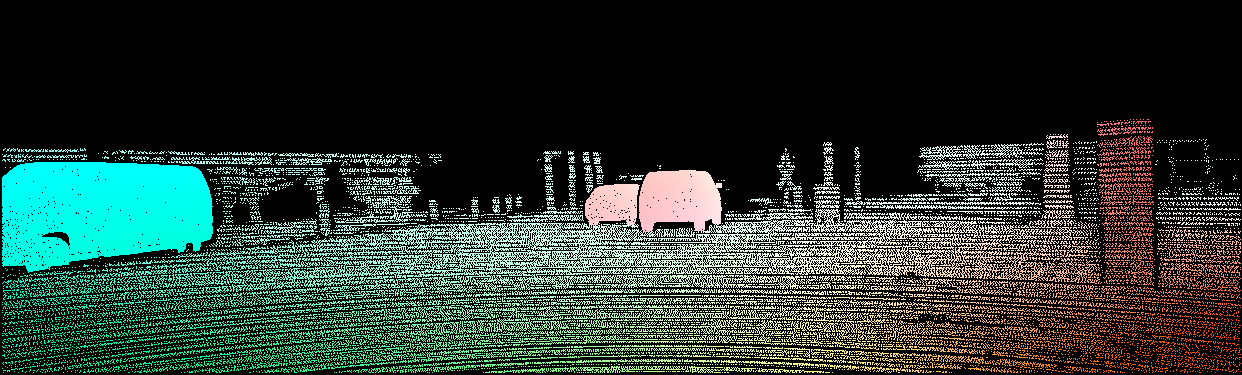} \\
 \includegraphics[width=1\textwidth]{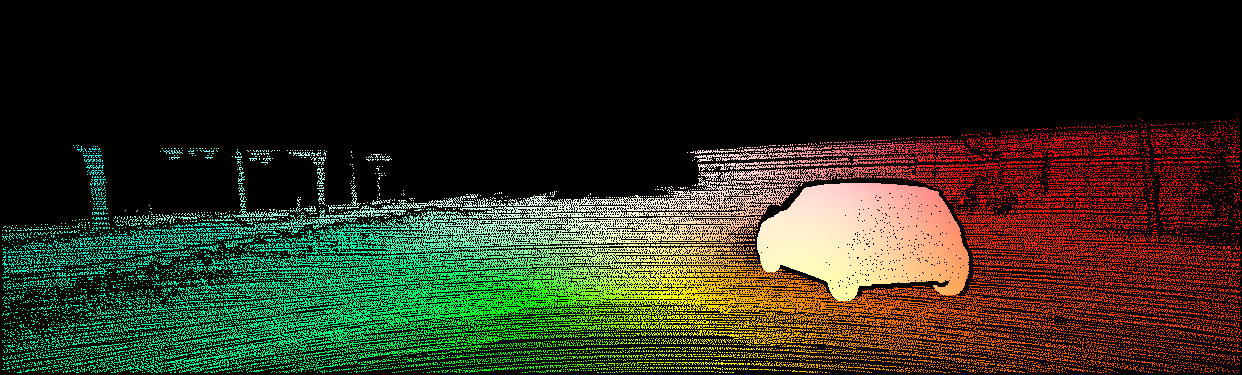} \\
 \includegraphics[width=1\textwidth]{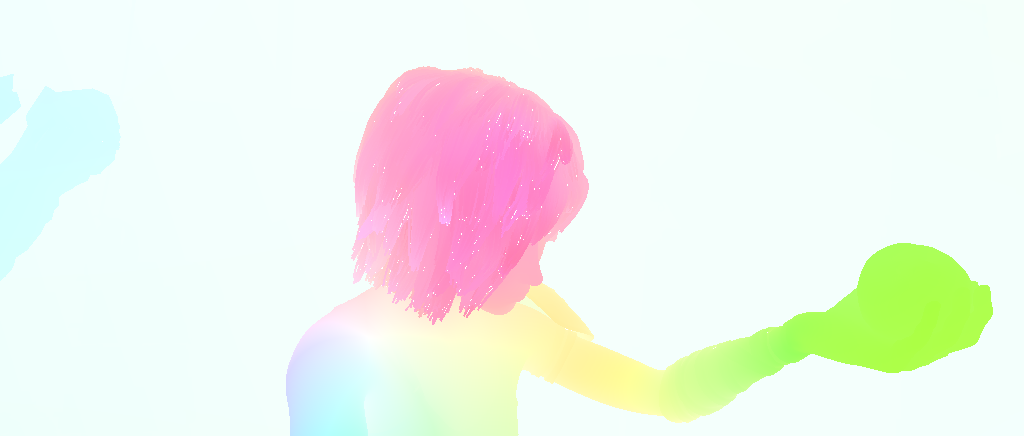} \\
 \includegraphics[width=1\textwidth]{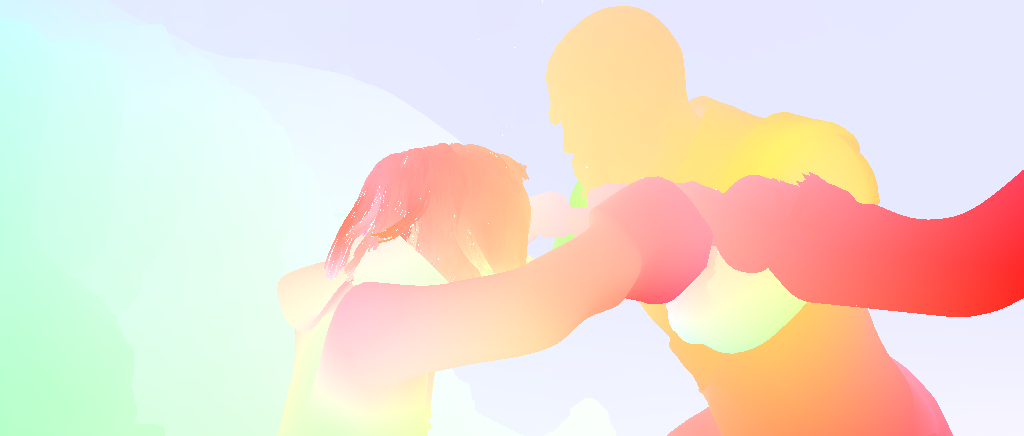}
 \end{minipage}
 }
 \hspace{-0.02\textwidth}
 \subfigure[DistillFlow Flow]{
 \begin{minipage}[t]{0.196\textwidth}
 \includegraphics[width=1\textwidth]{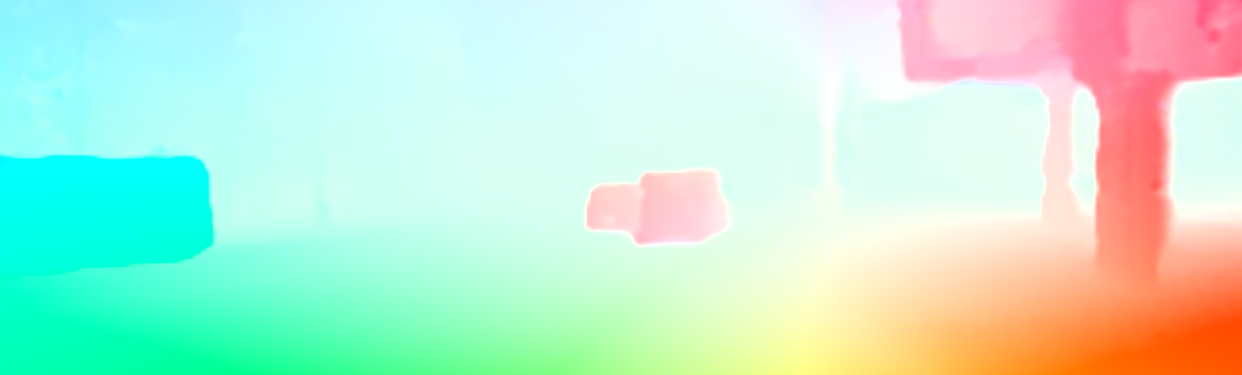} \\
 \includegraphics[width=1\textwidth]{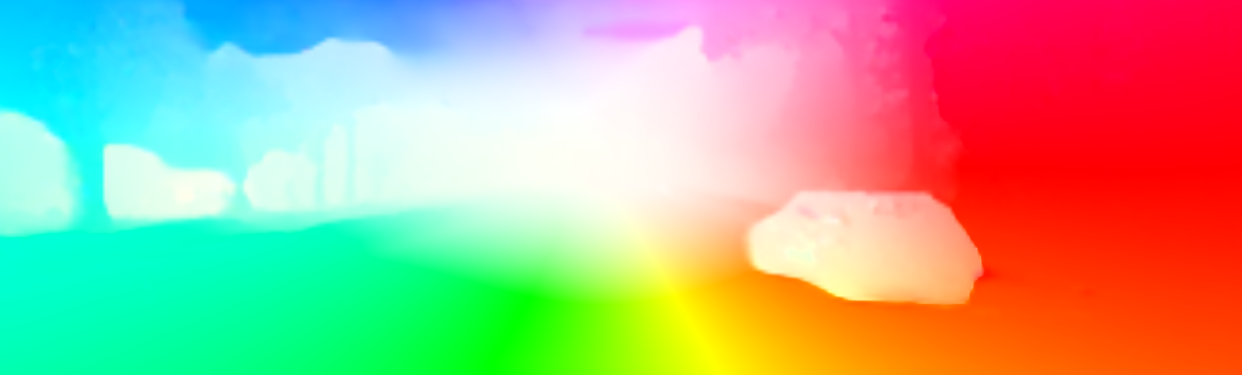} \\
 \includegraphics[width=1\textwidth]{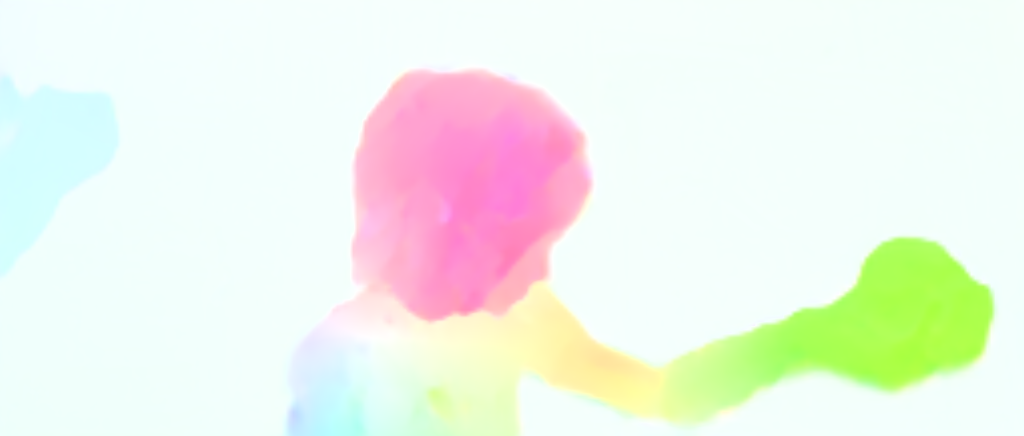} \\
 \includegraphics[width=1\textwidth]{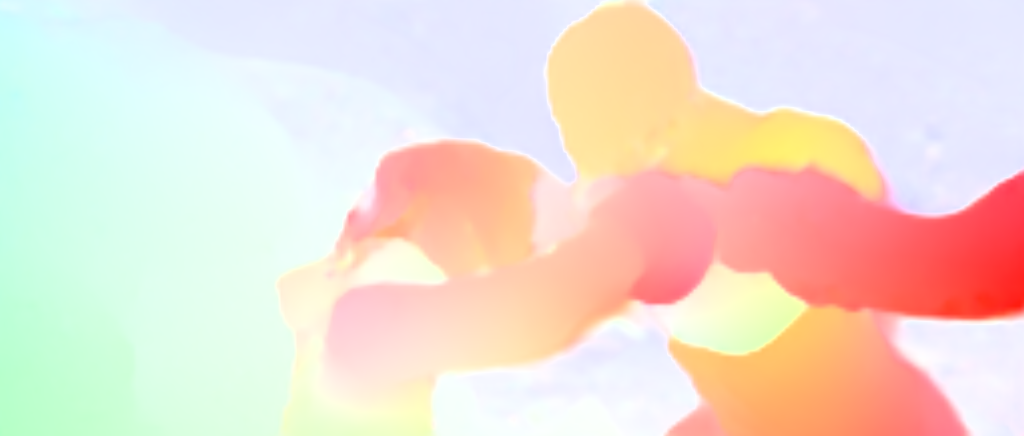}
 \end{minipage}
 }
 \hspace{-0.02\textwidth}
 \subfigure[GT Occlusion]{
 \begin{minipage}[t]{0.196\textwidth}
 \includegraphics[width=1\textwidth]{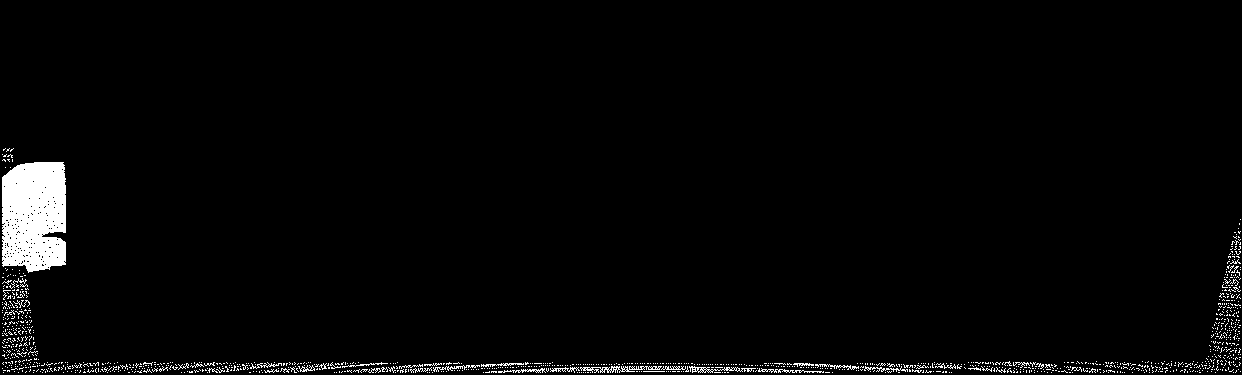} \\
 \includegraphics[width=1\textwidth]{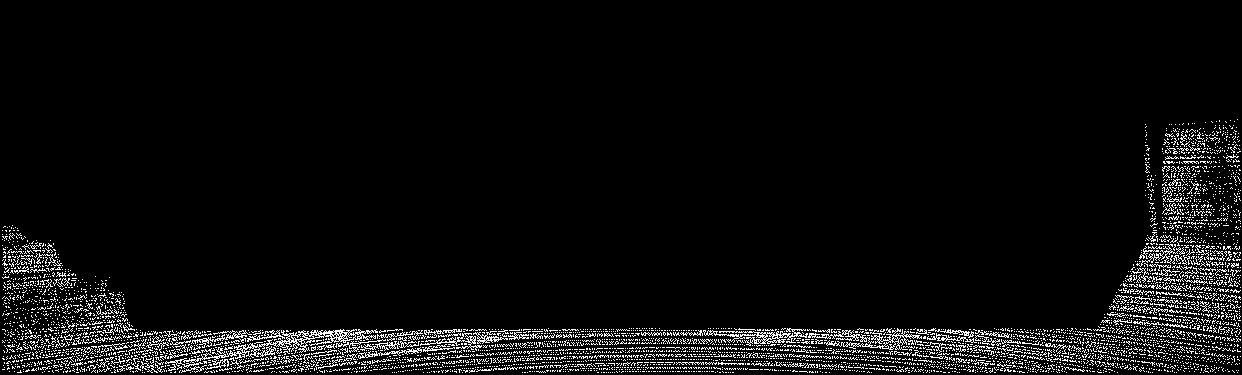} \\
 \includegraphics[width=1\textwidth]{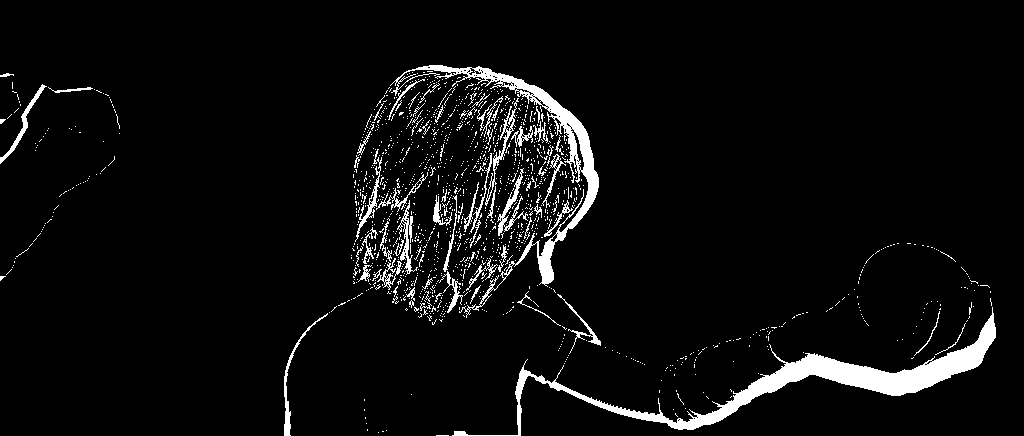} \\
 \includegraphics[width=1\textwidth]{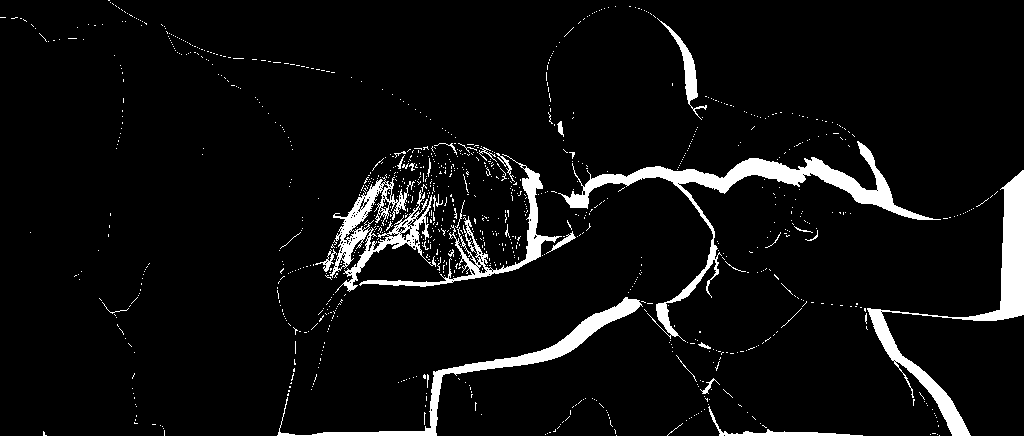}
 \end{minipage}
 }
 \hspace{-0.02\textwidth}
 \subfigure[DistillFlow Occlusion]{
 \begin{minipage}[t]{0.196\textwidth}
 \includegraphics[width=1\textwidth]{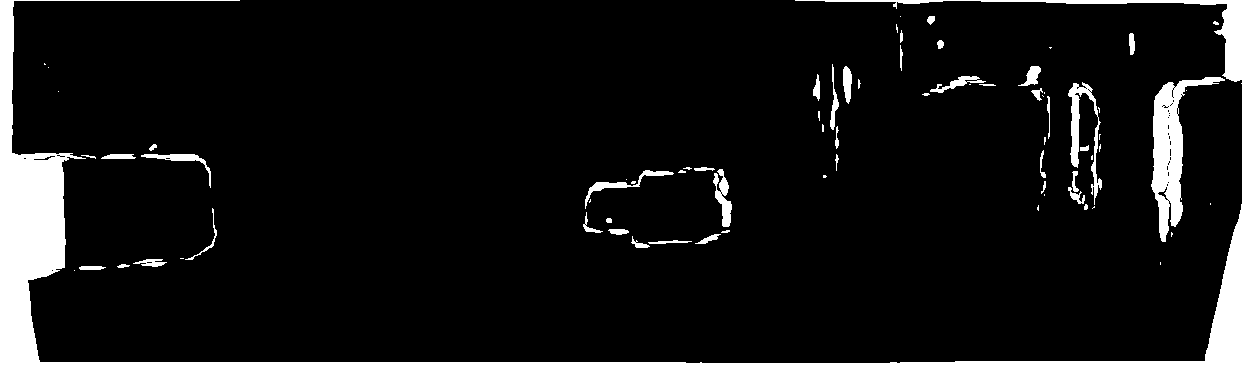} \\
 \includegraphics[width=1\textwidth]{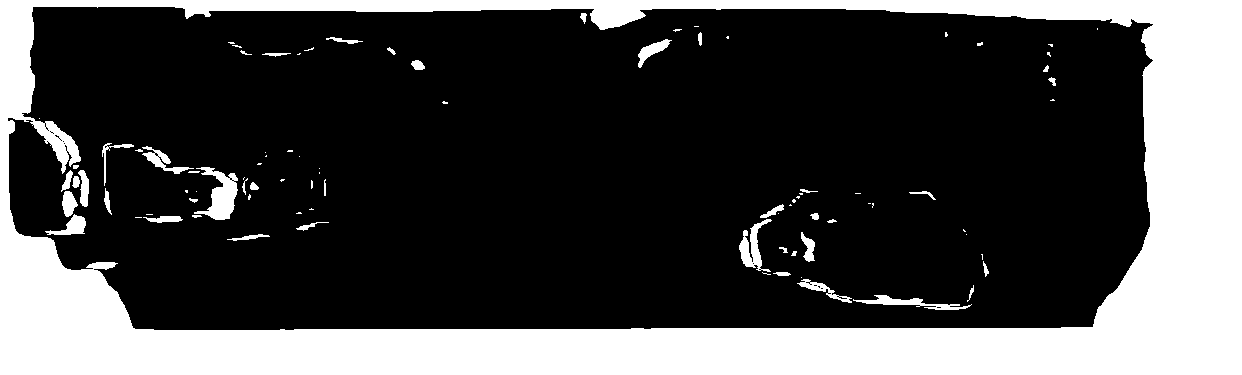} \\
 \includegraphics[width=1\textwidth]{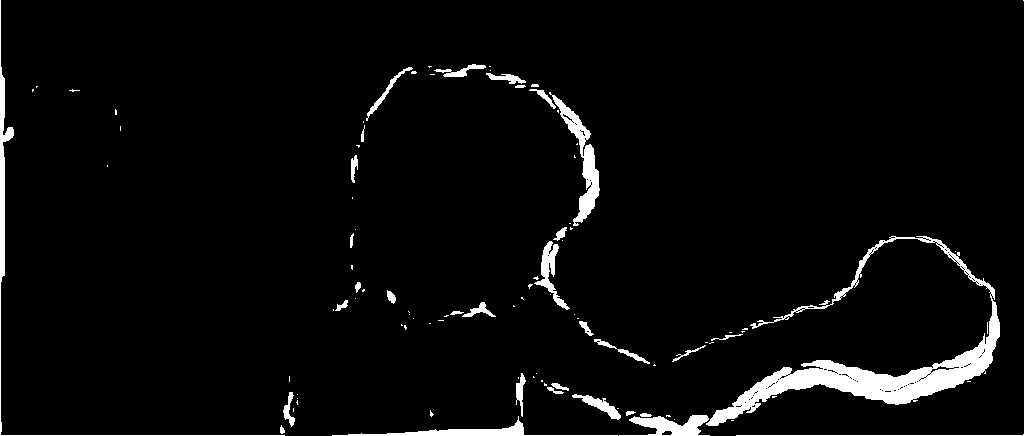} \\
 \includegraphics[width=1\textwidth]{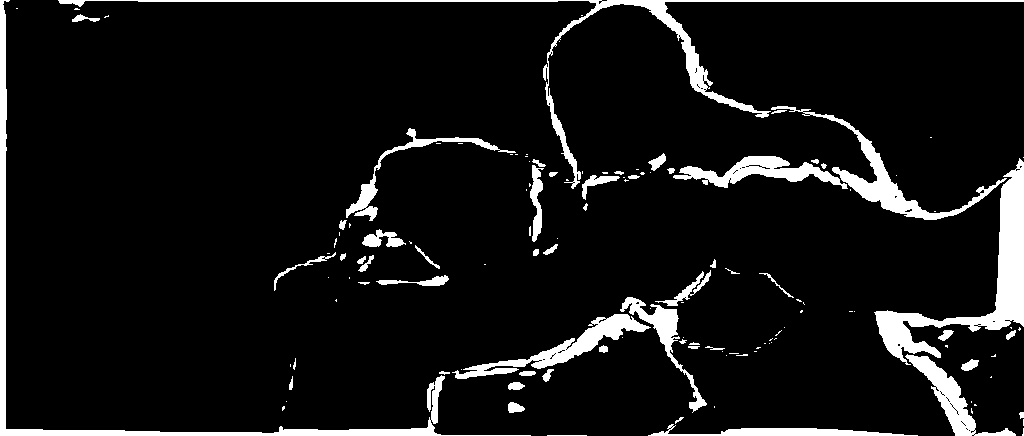}
 \end{minipage}
 }
 \hspace{-0.02\textwidth}
  \vspace{-1ex}
 \caption{Sample unsupervised results of DistillFlow on KITTI (top 2) and Sintel (bottom 2) training datasets. DistillFlow estimates both accurate optical flow and occlusion maps. Note that on KITTI datasets, the occlusion maps are sparse, which only contain pixels moving out of the image boundary.}
 \label{figure:sample_unsupervise}
 \vspace{-1ex}
 \end{figure*}
In our experiments, variant 1 and variant 2 achieve very similar performances. However, it takes more time to train variant 1 than variant 2. This is because for variant 1, apart from $L_{occ}$, we still need to compute $L_{pho}$, which is time-consuming. While for variant 2, we directly train it in a supervised manner after pre-computing pseudo flow and confidence maps. As a result, we set variant 2 as our default knowledge distillation strategy. However, its performance is limited by the prediction of the teacher model. To obtain more reliable predictions from stage 1, we employ model distillation to ensemble flow predictions from multiple teacher models.

In variant 2, challenging transformations create less confident predictions for both occluded and non-occluded pixels, therefore knowledge distillation improves the flow learning of both. However, variant 1 employs $L_{occ}$ and $L_{pho}$ to optimize occluded and non-occluded pixels individually. It is surprising how variant 1 achieves performance improvement over non-occluded pixels. We suspect the main reason is that forward-backward consistency check cannot accurately detect whether the pixel is occluded or not (\eg some non-occluded pixels are regarded as occluded by forward-backward consistency check). In other words, distillation loss $L_{occ}$ still optimizes a part of non-occluded pixels.

\subsection{Supervised Fine-tuning}
After pre-training on unlabeled datasets, we use real-world annotated data for fine-tuning. Since there are only annotations for forward flow $\textbf{w}_{f}$, we do not swap input image pairs. Suppose that the ground truth flow is $\textbf{w}^{gt}_{f}$, and mask $V$ denotes whether the pixel has a label, where value 1 means that the pixel has a valid ground truth flow.  Then we can obtain the supervised fine-tuning loss as follow:
\begin{equation}
L_{sup} = \sum(\psi(\textbf{w}^{gt}_{f} - \textbf{w}_{f}) \odot V) / \sum{V}.
\end{equation}
During fine-tuning, We first initialize the model with the self-supervised pre-trained student on each dataset, then optimize it using $L_{sup}$. Inspired by the knowledge distillation from unsupervised flow learning, we extend the idea of distillation to semi-supervised learning. That is, after supervised fine-tuning, we can compute reliable flow and confidence maps for unlabeled data, which are denoted as self-annotated data. Then we mix the real annotated data and self-annotated data and train our model in a supervised manner. Note that the count of real annotated data is very limited, therefore we make a balance between real annotated data and self-annotated data. Suppose there are $n_1$ real annotated image pairs, $n_2$ self-annotated image pairs, we will repeat real annotated image pairs $\frac {n_2}  {n_1}$ times.

\begin{figure*}[t]
 \centering
 \subfigure[Input Images]{
 \begin{minipage}[t]{0.196\textwidth}
 \begin{overpic} [width=1\textwidth]{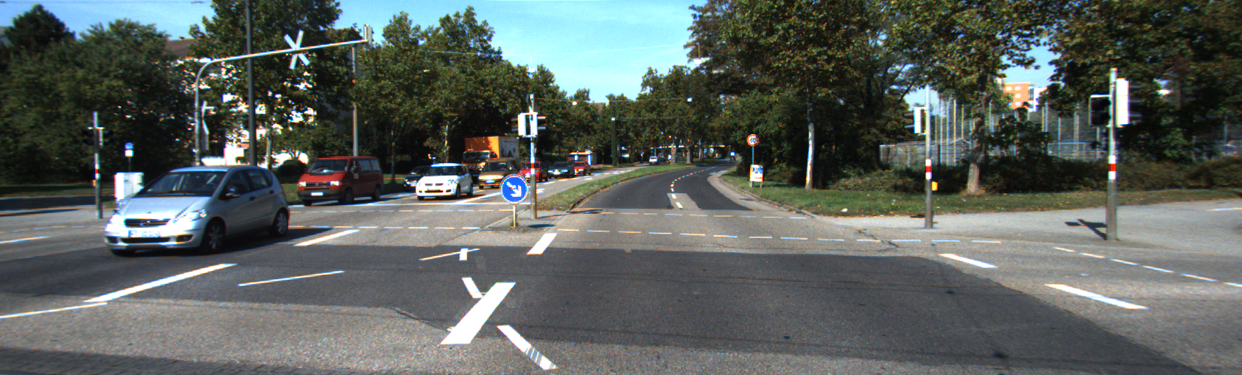} \put(4, 20){\color{red} $I_1$} \end{overpic} \\
 \begin{overpic} [width=1\textwidth]{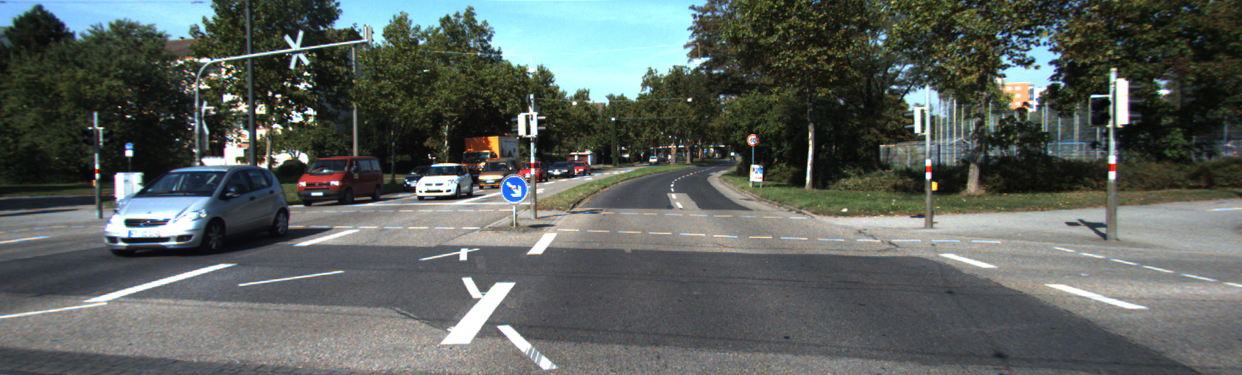} \put(4, 20){\color{red} $I_2$} \end{overpic} \\
 \begin{overpic} [width=1\textwidth]{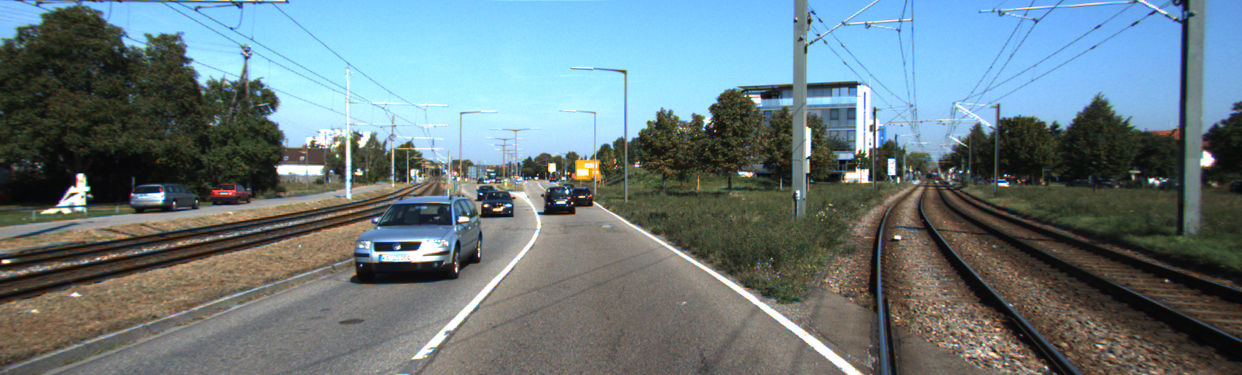} \put(4, 20){\color{red} $I_1$} \end{overpic} \\
 \begin{overpic} [width=1\textwidth]{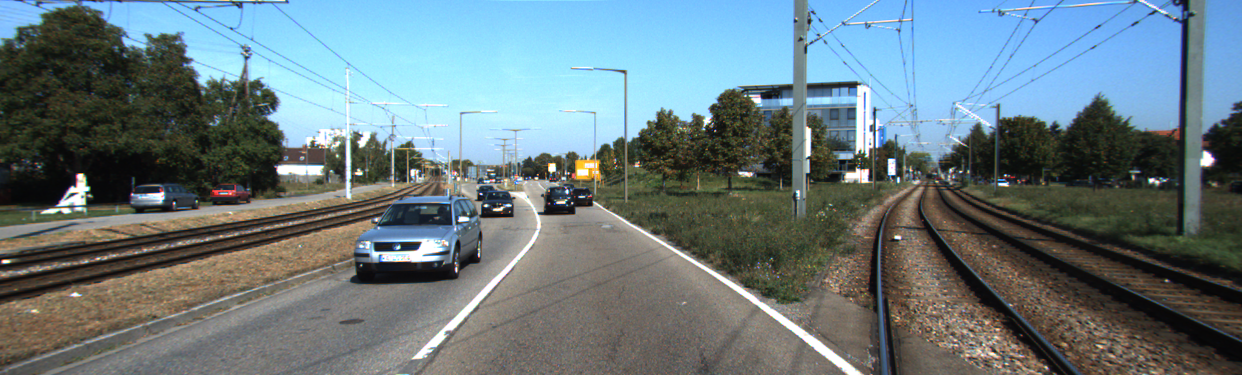} \put(4, 20){\color{red} $I_2$} \end{overpic}
 \end{minipage}
 }
 \hspace{-0.02\textwidth}
 \subfigure[Back2FutureFlow~\cite{Janai2018ECCV}]{
 \begin{minipage}[t]{0.196\textwidth}
 \begin{overpic} [width=1\textwidth]{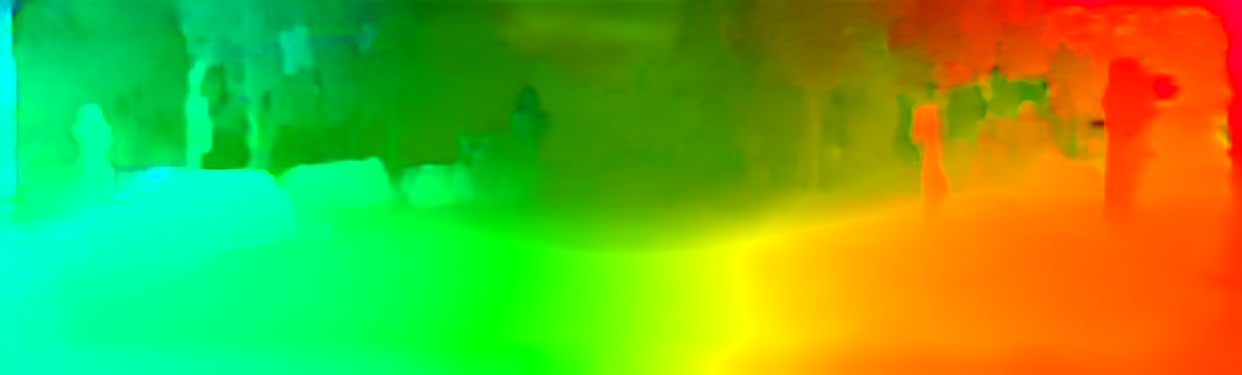} \put(4, 20){\color{white} Fl: 22.65\%} \end{overpic} \\
 \begin{overpic} [width=1\textwidth]{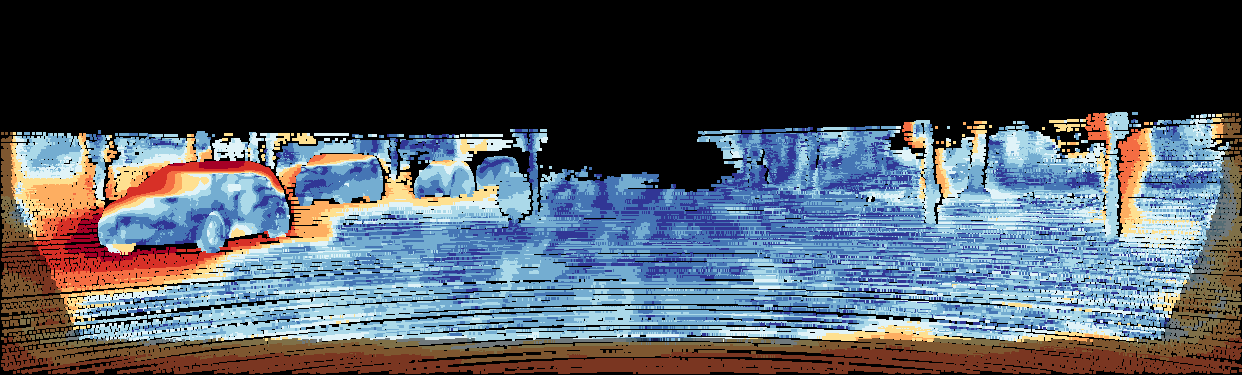} \end{overpic}\\
 \begin{overpic} [width=1\textwidth]{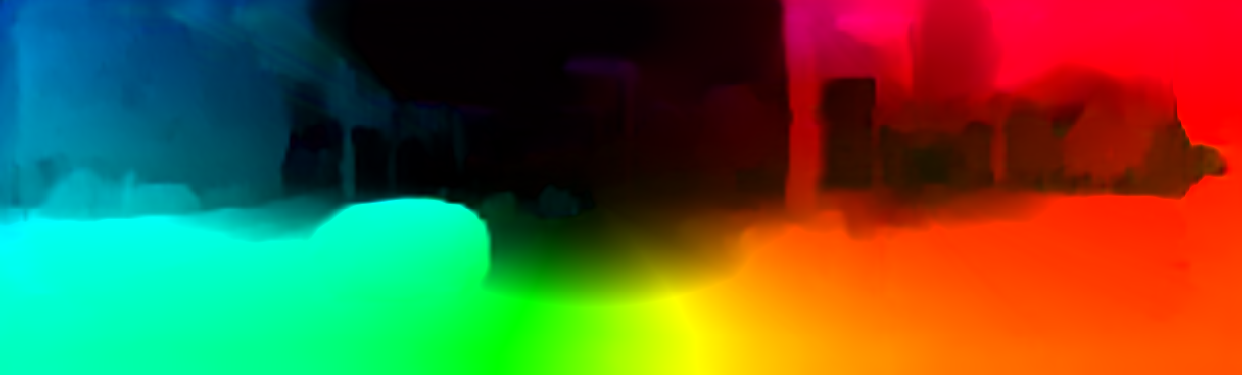} \put(4, 20){\color{white} Fl: 27.14\%} \end{overpic} \\
 \begin{overpic} [width=1\textwidth]{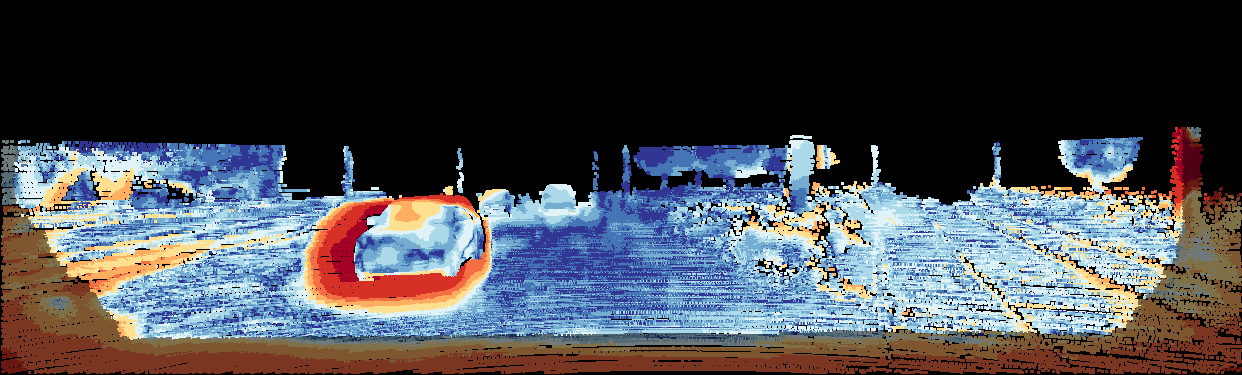} \end{overpic}
 \end{minipage}
 }
 \hspace{-0.02\textwidth}
 \subfigure[SelFlow~\cite{Liu:2019:SelFlow}]{
 \begin{minipage}[t]{0.196\textwidth}
 \begin{overpic} [width=1\textwidth]{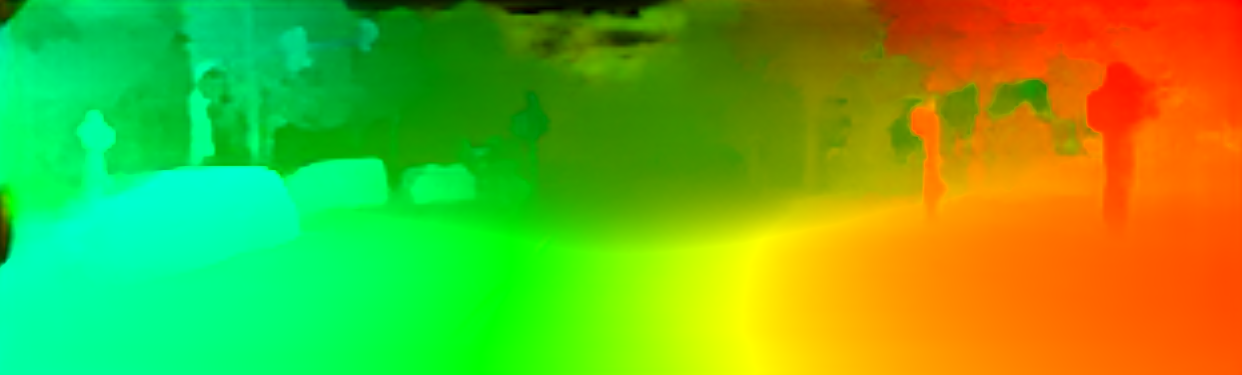} \put(4, 20){\color{white} Fl: 9.17\%} \end{overpic} \\
 \begin{overpic} [width=1\textwidth]{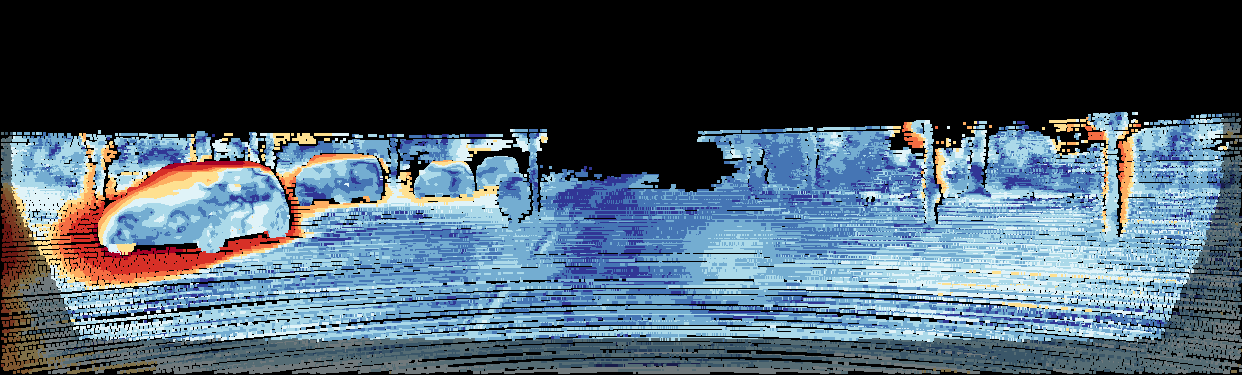} \end{overpic}\\
 \begin{overpic} [width=1\textwidth]{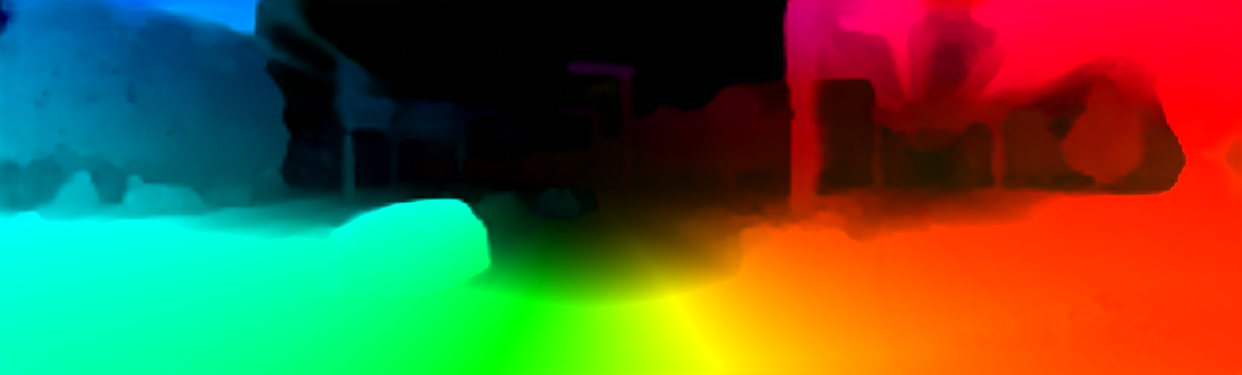} \put(4, 20){\color{white} Fl: 16.40\%} \end{overpic} \\
 \begin{overpic} [width=1\textwidth]{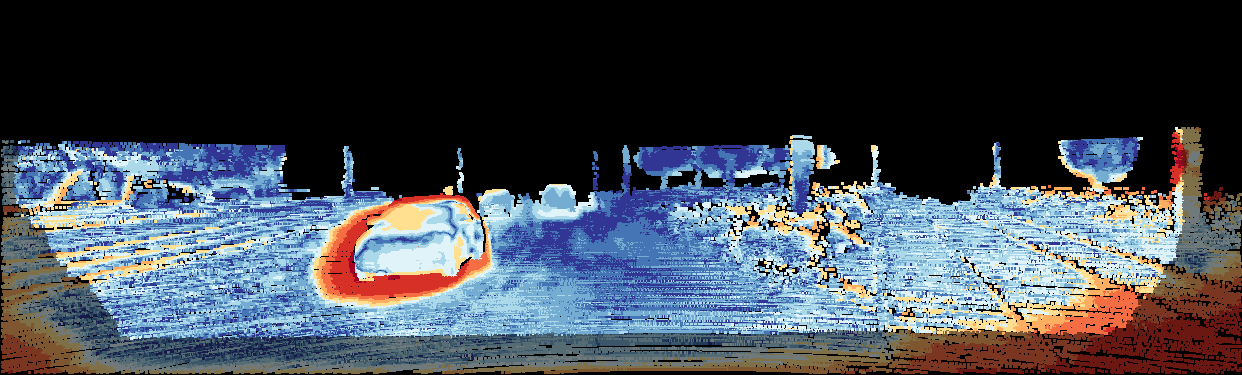} \end{overpic}
 \end{minipage}
 }
 \hspace{-0.02\textwidth}
 \subfigure[Flow2Stereo~\cite{Liu:2020:Flow2Stereo}]{
 \begin{minipage}[t]{0.196\textwidth}
 \begin{overpic} [width=1\textwidth]{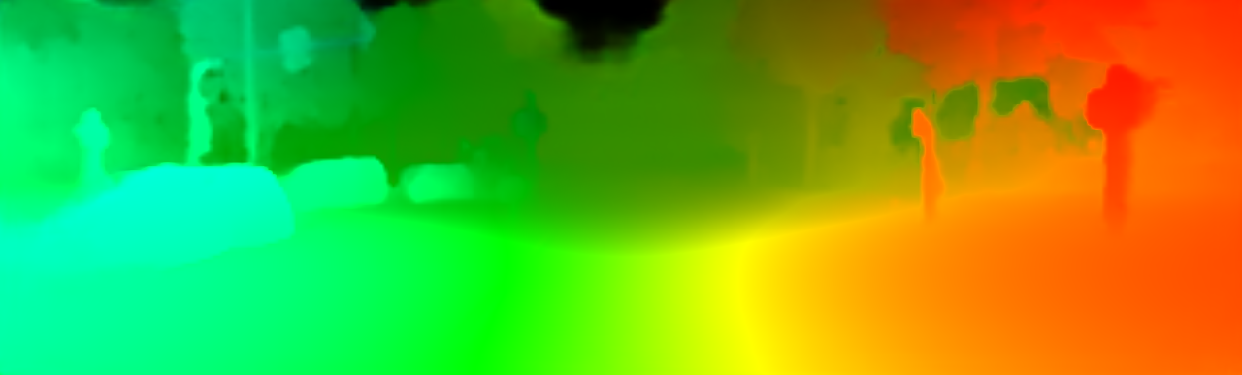} \put(4, 20){\color{white} Fl: 7.41\%} \end{overpic} \\
 \begin{overpic} [width=1\textwidth]{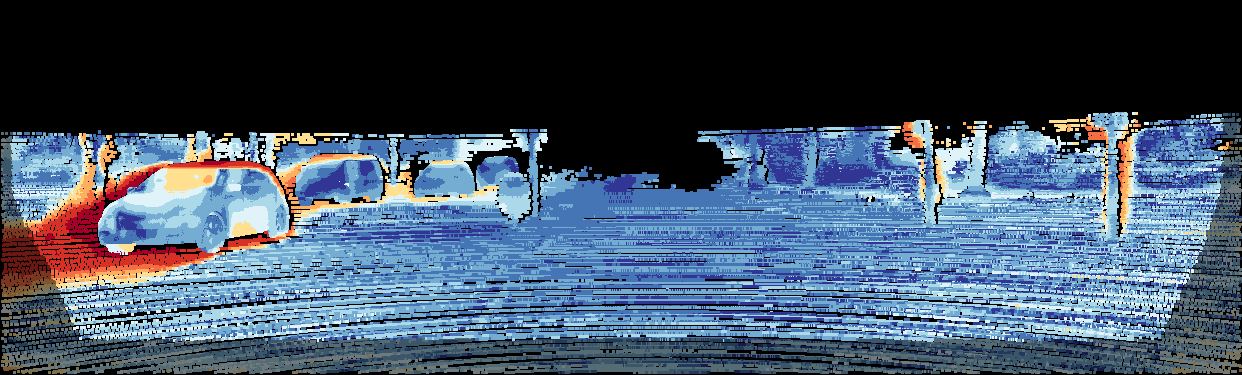} \end{overpic}\\
 \begin{overpic} [width=1\textwidth]{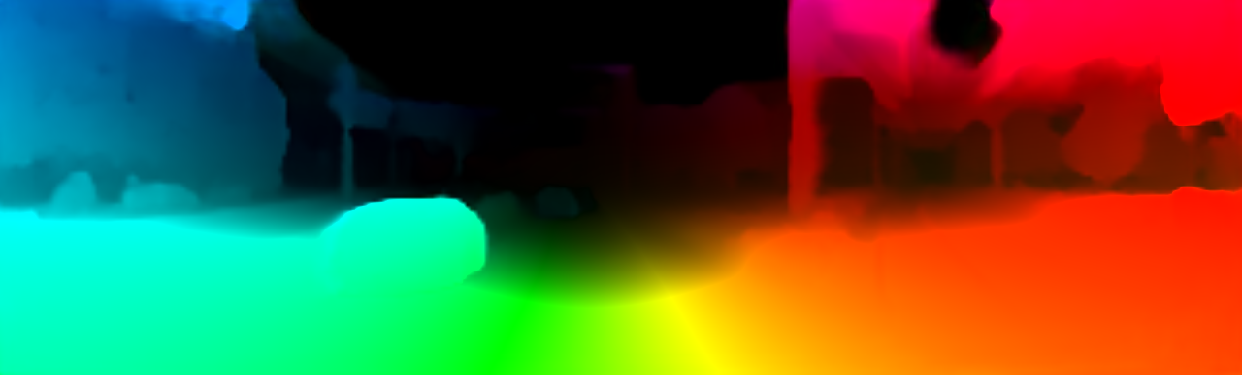} \put(4, 20){\color{white} Fl: 9.59\%} \end{overpic} \\
 \begin{overpic} [width=1\textwidth]{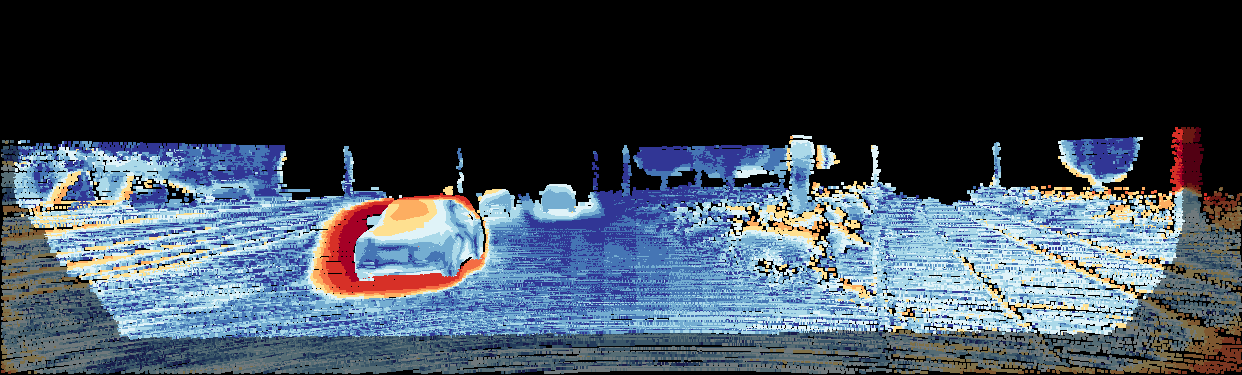} \end{overpic}
 \end{minipage}
 }
 \hspace{-0.02\textwidth}
 \subfigure[DistillFlow]{
 \begin{minipage}[t]{0.196\textwidth}
 \begin{overpic} [width=1\textwidth]{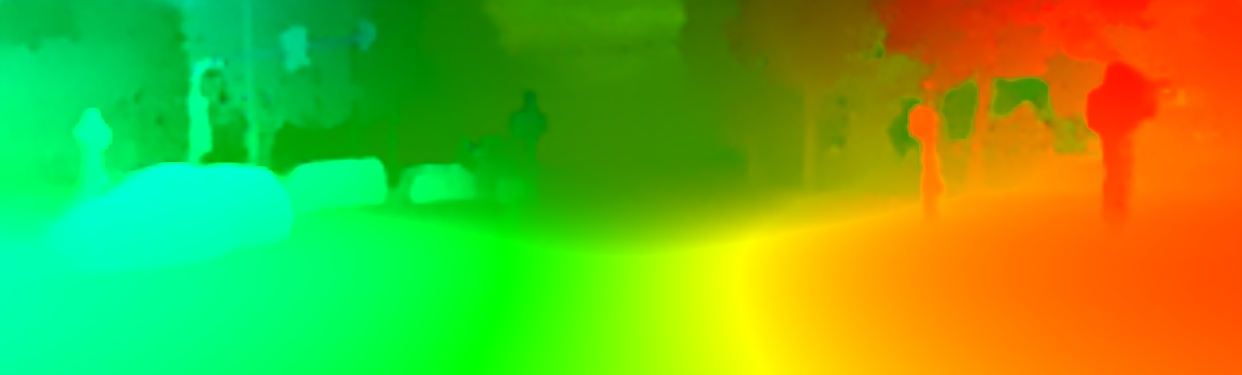} \put(4, 20){\color{white} Fl: 6.63\%} \end{overpic} \\
 \begin{overpic} [width=1\textwidth]{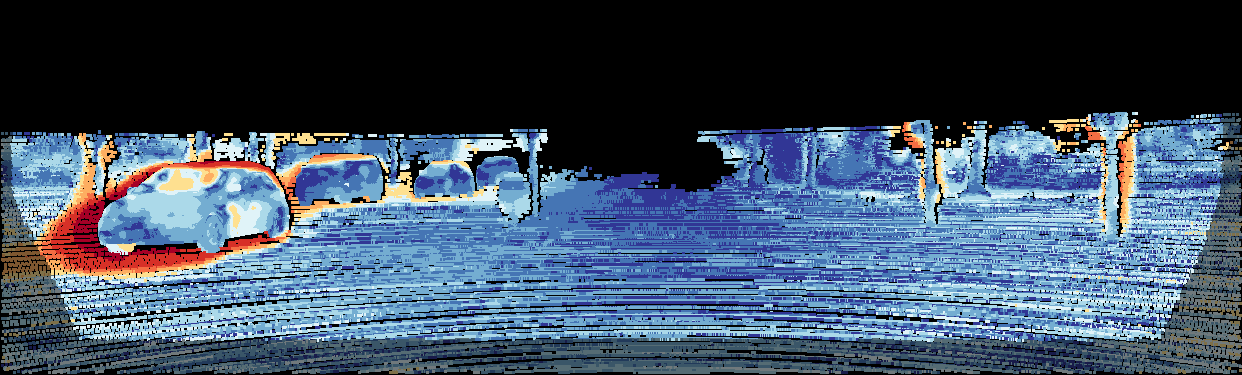} \end{overpic}\\
 \begin{overpic} [width=1\textwidth]{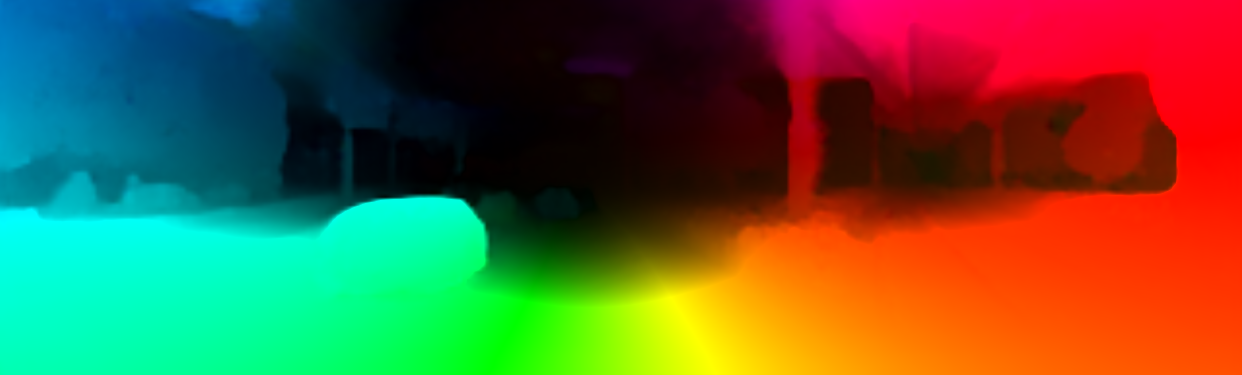} \put(4, 20){\color{white} Fl: 8.04\%} \end{overpic} \\
 \begin{overpic} [width=1\textwidth]{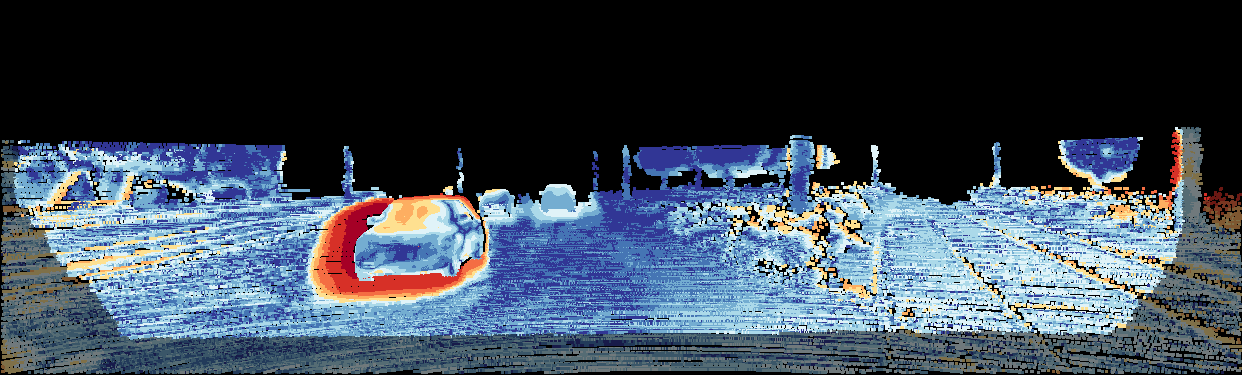} \end{overpic}
 \end{minipage}
 }
 \hspace{-0.02\textwidth}
 \vspace{-1ex}
 \caption{Qualitative comparison with state-of-the-art unsupervised learning methods on the KITTI 2015 benchmark. For each case, the top row is optical flow and the bottom row is error map. The redder the color in the error map, the greater the error. More examples are available on KITTI 2015 benchmark.}
  \vspace{-1ex}
 \label{figure:benchmark_kitti_unsupervise}
 \end{figure*}

\begin{figure*}[t]
 \centering
 \subfigure[Input Images]{
 \begin{minipage}[t]{0.196\textwidth}
 \begin{overpic} [width=1\textwidth]{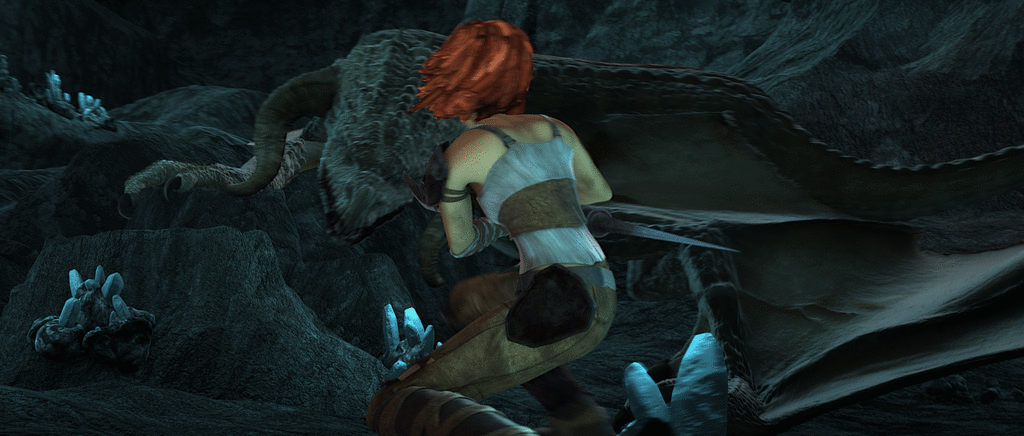} \put(4, 32){\color{red} $I_1$} \end{overpic} \\
 \begin{overpic} [width=1\textwidth]{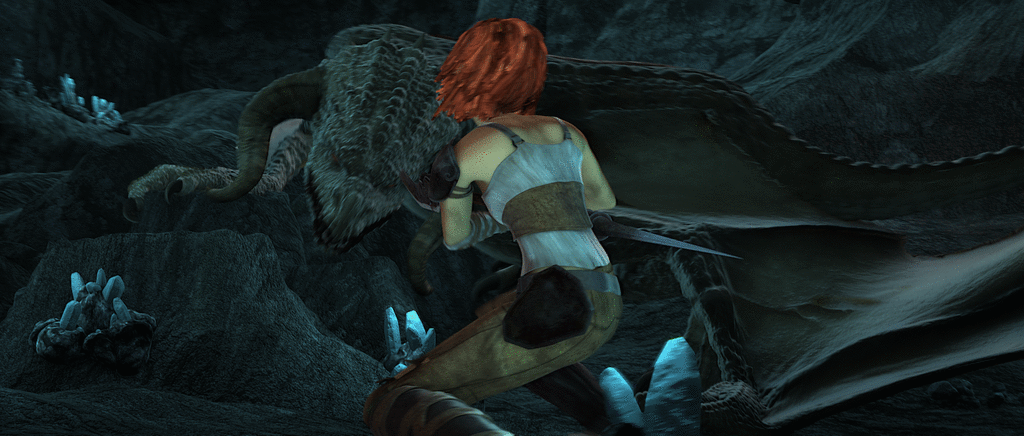} \put(4, 32){\color{red} $I_2$} \end{overpic} \\
 \begin{overpic} [width=1\textwidth]{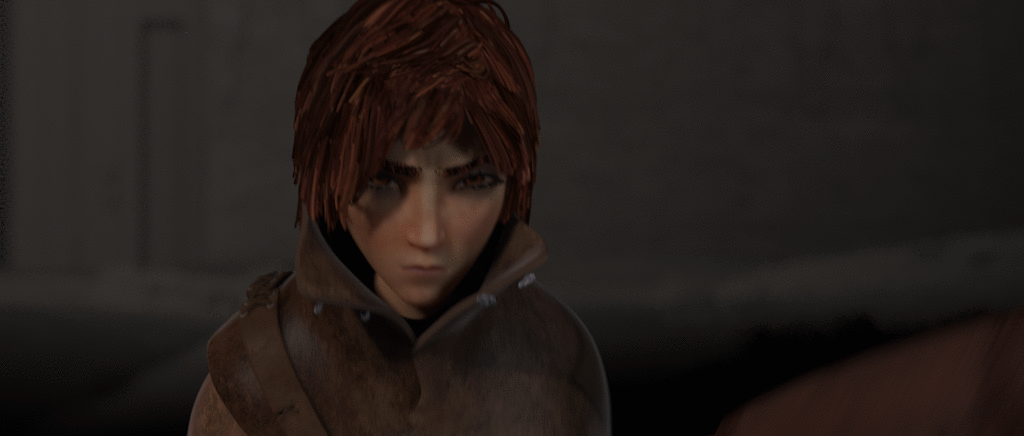} \put(4, 32){\color{red} $I_1$} \end{overpic} \\
 \begin{overpic} [width=1\textwidth]{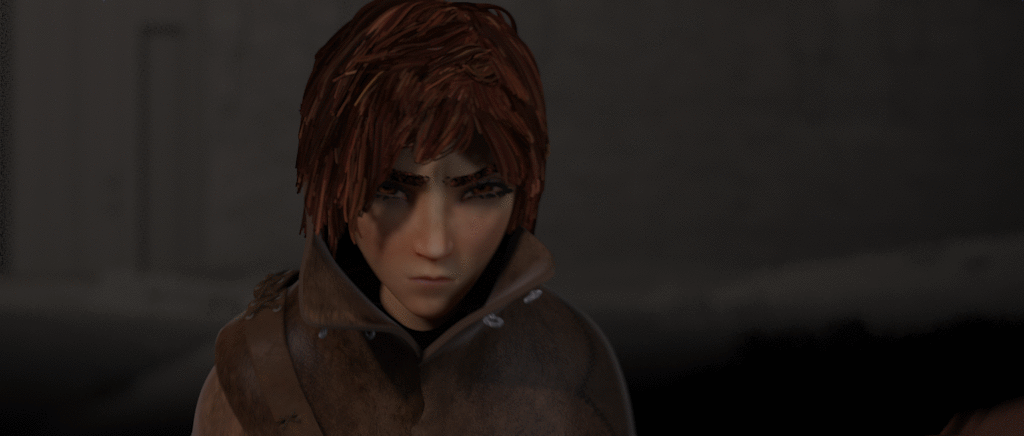} \put(4, 32){\color{red} $I_2$} \end{overpic}
 \end{minipage}
 }
 \hspace{-0.02\textwidth}
 \subfigure[Back2FutureFlow~\cite{Janai2018ECCV}]{
 \begin{minipage}[t]{0.196\textwidth}
 \begin{overpic} [width=1\textwidth]{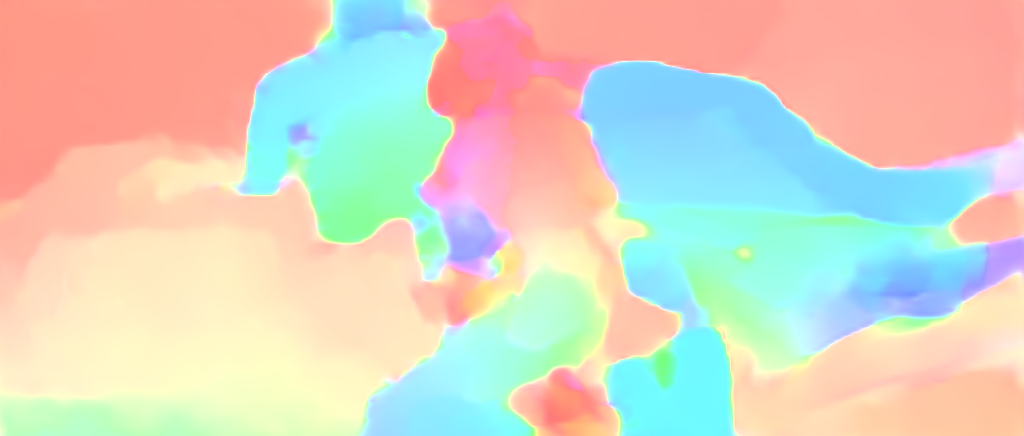} \put(4, 32){\color{black} EPE: 9.313} \end{overpic} \\
 \begin{overpic} [width=1\textwidth]{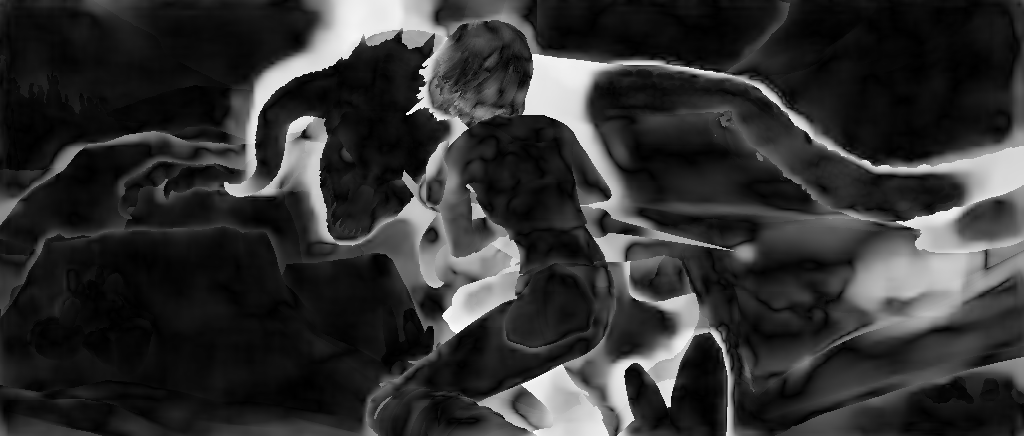} \end{overpic}\\
 \begin{overpic} [width=1\textwidth]{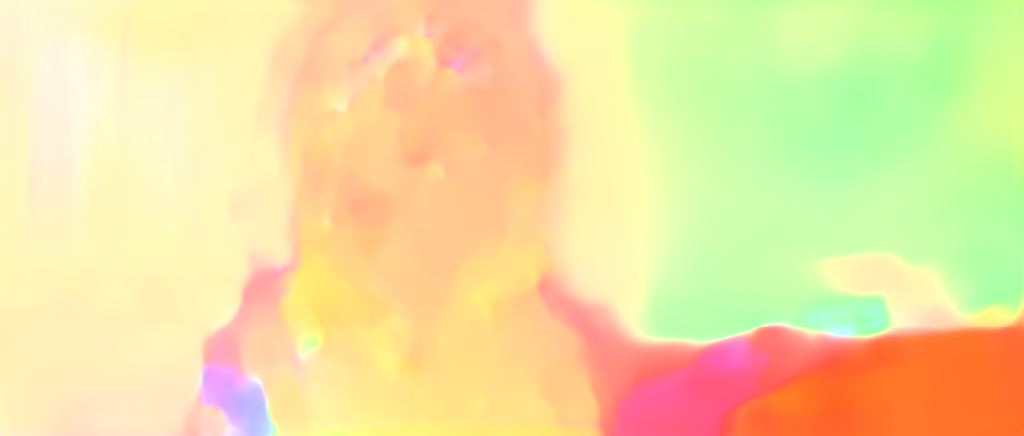} \put(4, 32){\color{black} EPE: 8.690} \end{overpic} \\
 \begin{overpic} [width=1\textwidth]{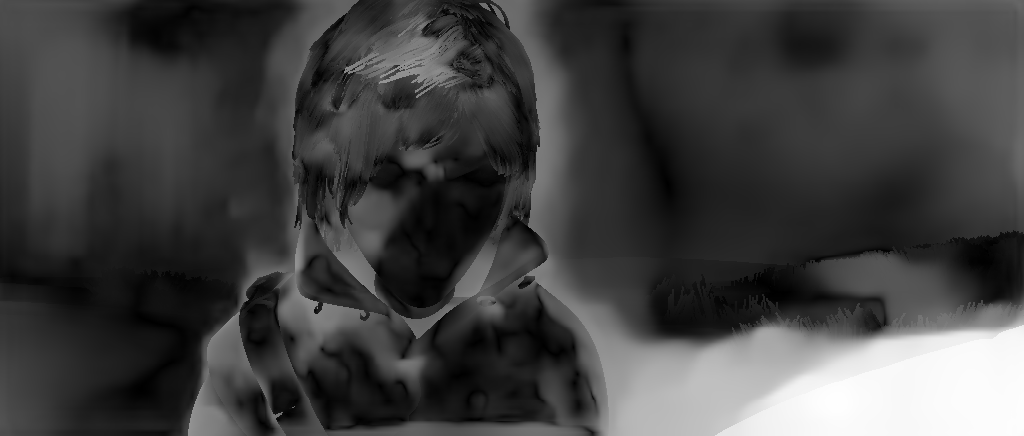} \end{overpic}
 \end{minipage}
 }
 \hspace{-0.02\textwidth}
 \subfigure[EpipolarFlow~\cite{zhong2019epiflow}]{
 \begin{minipage}[t]{0.196\textwidth}
 \begin{overpic} [width=1\textwidth]{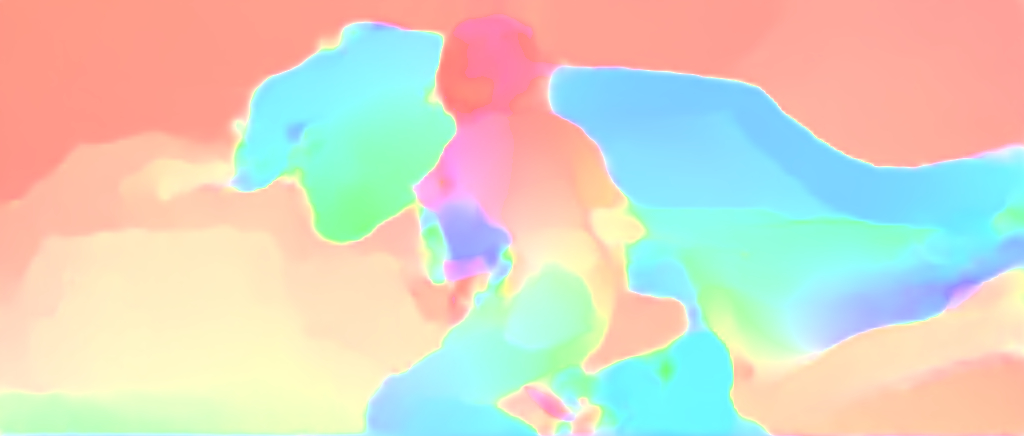} \put(4, 32){\color{black} EPE: 9.198} \end{overpic} \\
 \begin{overpic} [width=1\textwidth]{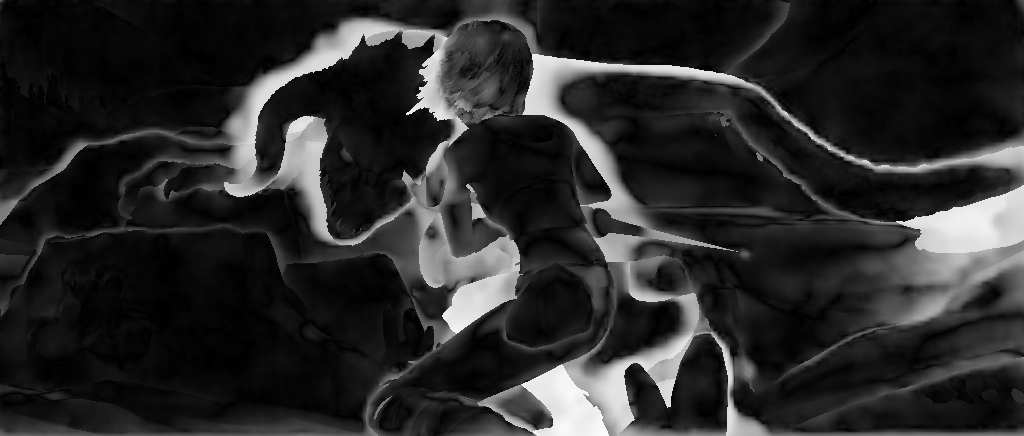} \end{overpic}\\
 \begin{overpic} [width=1\textwidth]{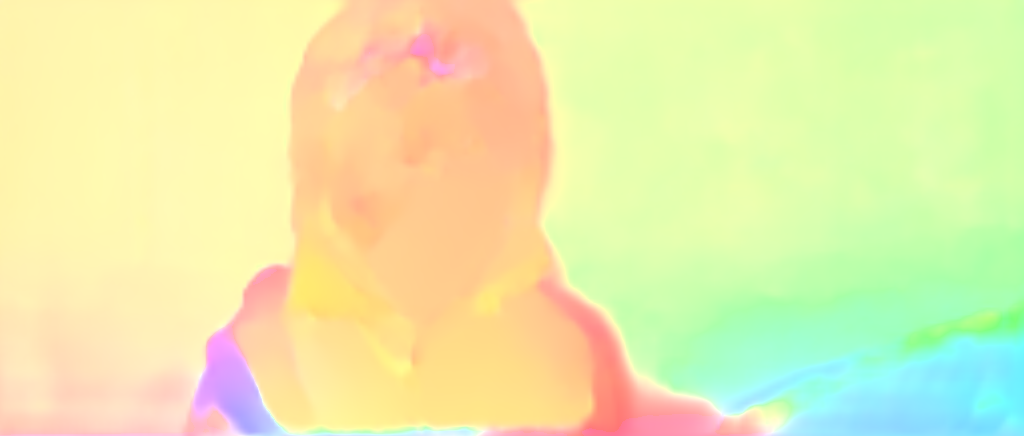} \put(4, 32){\color{black} EPE: 6.479} \end{overpic} \\
 \begin{overpic} [width=1\textwidth]{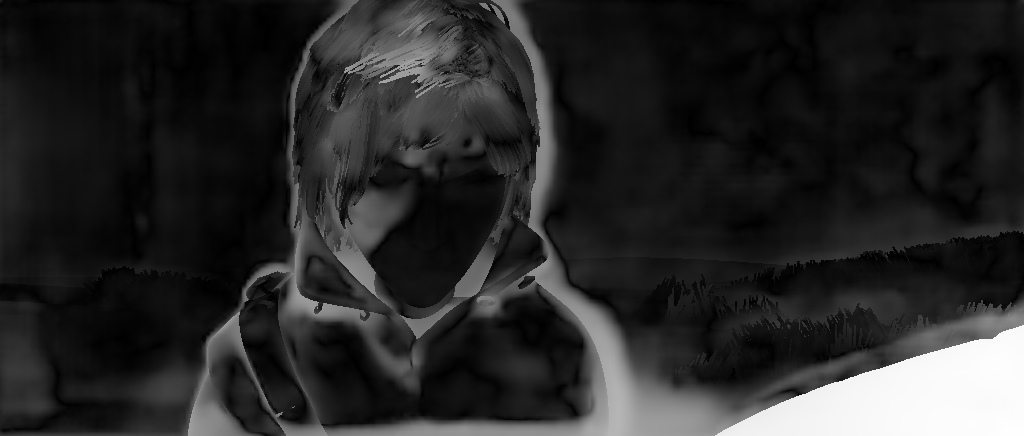} \end{overpic}
 \end{minipage}
 }
 \hspace{-0.02\textwidth}
 \subfigure[SelFlow~\cite{Liu:2019:SelFlow}]{
 \begin{minipage}[t]{0.196\textwidth}
 \begin{overpic} [width=1\textwidth]{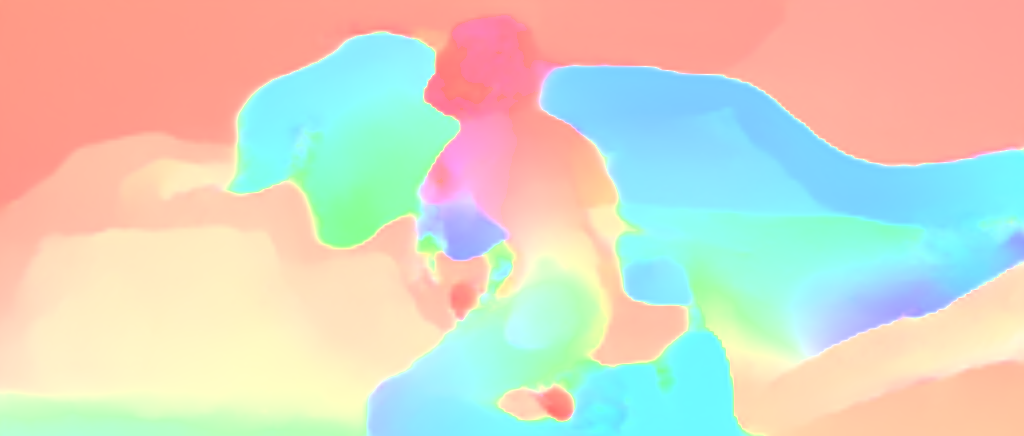} \put(4, 32){\color{black} EPE: 7.168} \end{overpic} \\
 \begin{overpic} [width=1\textwidth]{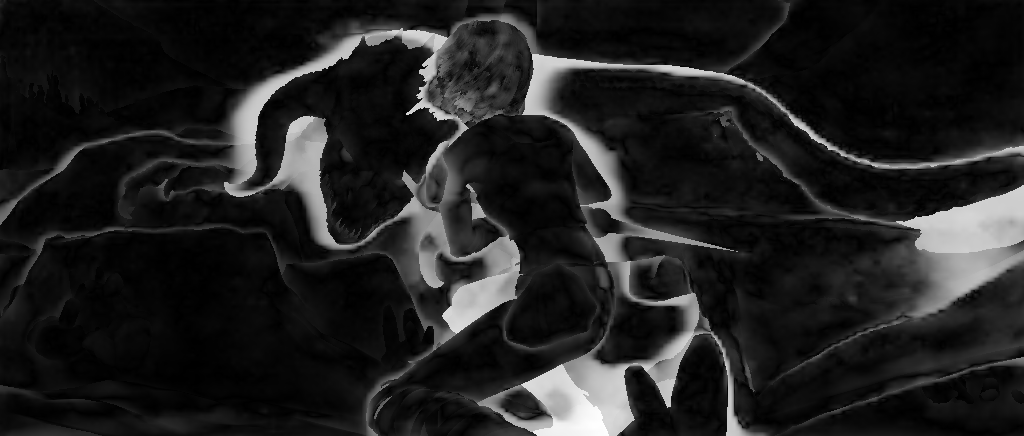} \end{overpic}\\
 \begin{overpic} [width=1\textwidth]{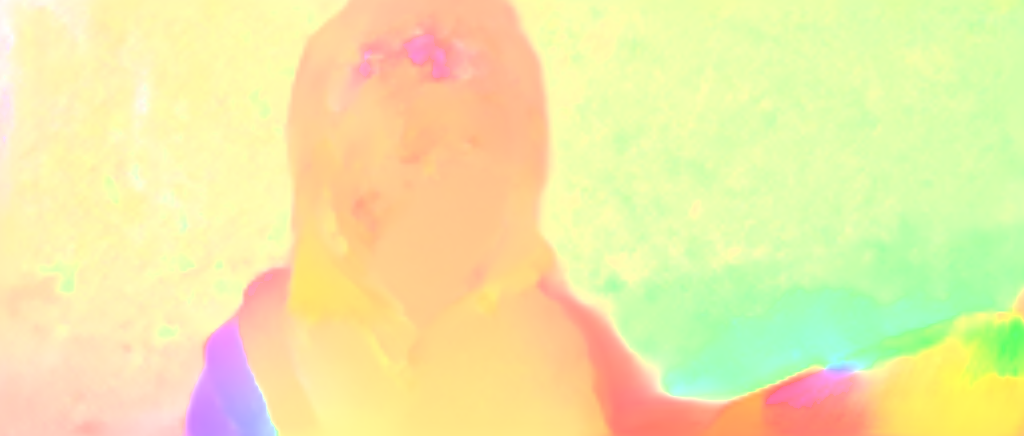} \put(4, 32){\color{black} EPE: 6.081} \end{overpic} \\
 \begin{overpic} [width=1\textwidth]{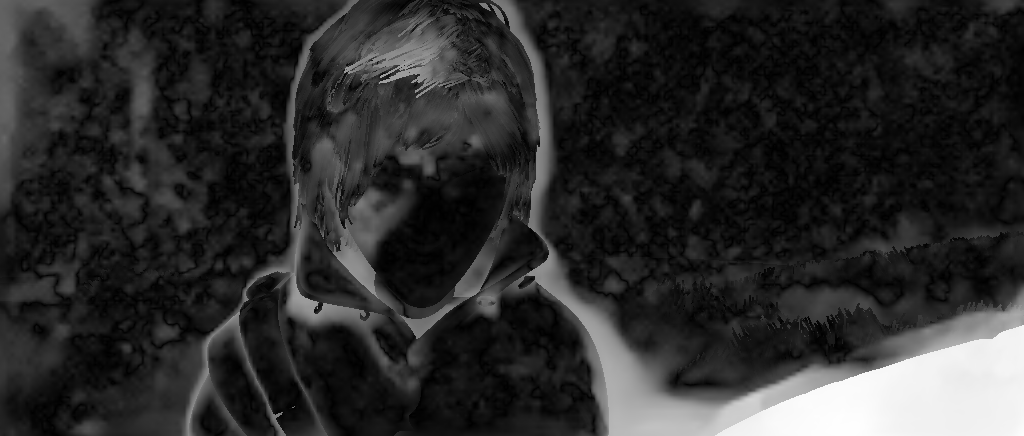} \end{overpic}
 \end{minipage}
 }
 \hspace{-0.02\textwidth}
 \subfigure[DistillFlow]{
 \begin{minipage}[t]{0.196\textwidth}
 \begin{overpic} [width=1\textwidth]{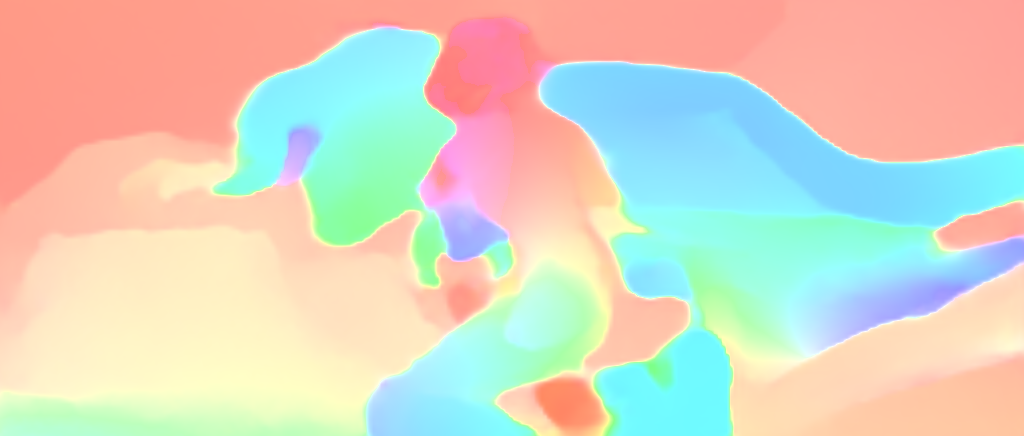} \put(4, 32){\color{black} EPE: 6.749} \end{overpic} \\
 \begin{overpic} [width=1\textwidth]{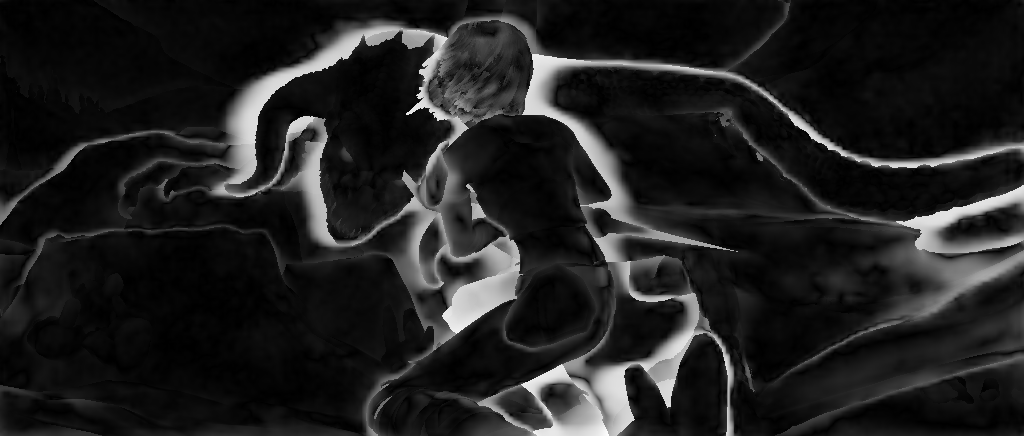} \end{overpic}\\
 \begin{overpic} [width=1\textwidth]{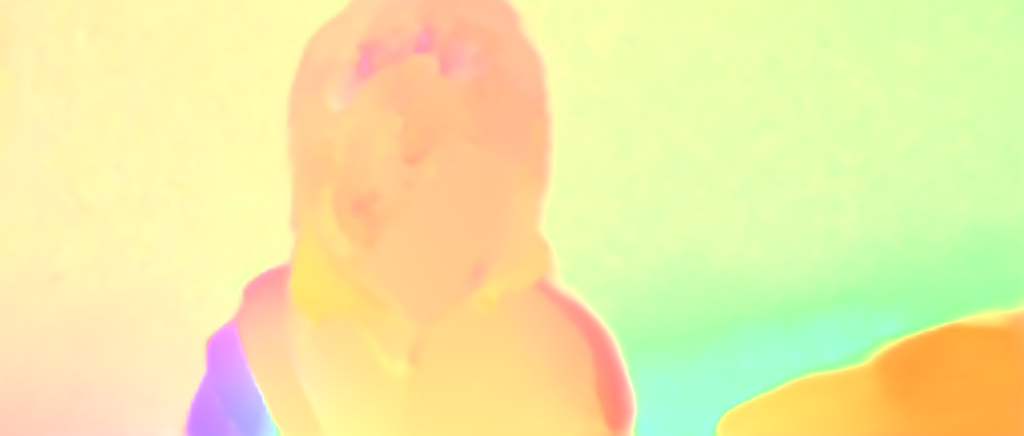} \put(4, 32){\color{black} EPE: 5.786} \end{overpic} \\
 \begin{overpic} [width=1\textwidth]{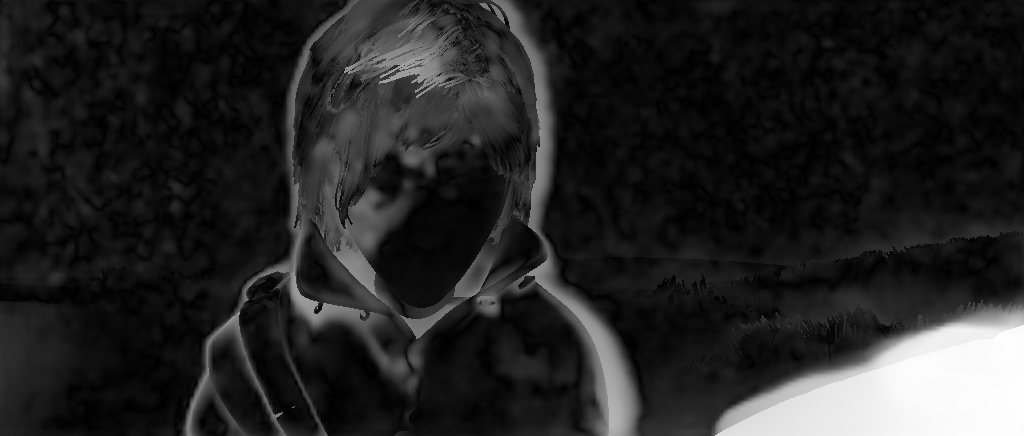} \end{overpic}
 \end{minipage}
 }
 \hspace{-0.02\textwidth}
 \vspace{-1ex}
 \caption{Qualitative comparison with state-of-the-art unsupervised learning methods on Sintel Final benchmark. For each case, the top row is optical flow and the bottom row is error map. The whiter the color in the error map, the greater the error. More examples are available on the Sintel benchmark.}
 \label{figure:benchmark_sintel_unsupervise}
 \vspace{-1ex}
 \end{figure*}

\begin{table*}[ht]
\caption{Quantitative evaluation of optical flow estimation on KITTI 2012 and KITTI 2015 datasets. Missing entries (-) indicate that the results are not reported for the respective method. Bold fonts highlight the best results among unsupervised and supervised methods. Parentheses mean that training and testing are performed on the same dataset. \texttt{fg} and \texttt{bg} denote results of foreground and background regions respectively. \texttt{(+Stereo)} denotes stereo data is used during training, and \texttt{*} denotes using more than two frames to estimate flow.}
\label{table:main_kitti_flow}
\centering
\resizebox{1\textwidth}{!}{
\begin{tabular}{clccccccccccc}
   \toprule
   &   \multirow{3}{*}{Method} & \multicolumn{6}{c}{KITTI 2012} & \multicolumn{5}{c}{KITTI 2015} \\
   \cmidrule(l{3mm}r{3mm}){3-8}    \cmidrule(l{3mm}r{3mm}){9-13} 
   &  &\multicolumn{2}{c}{train} &\multicolumn{4}{c}{test} & \multicolumn{2}{c}{train} & \multicolumn{3}{c}{test}  \\
   \cmidrule(l{3mm}r{3mm}){3-4}  \cmidrule(l{3mm}r{3mm}){5-8} \cmidrule(l{3mm}r{3mm}){9-10} \cmidrule(l{3mm}r{3mm}){11-13}
   &  & EPE-all & EPE-noc & EPE-all & EPE-noc & Fl-all & Fl-noc & EPE-all & EPE-noc & Fl-all &Fl-fg & Fl-bg \\
   \midrule
   \multirow{14}{*}{\rotatebox[origin=c]{90}{Unsupervised}}
   & BackToBasic~\cite{jason2016back}             & 11.3  & 4.3  & 9.9  & 4.6  & 43.15\% & 34.85\% &  --   &  --   &  --      &    --   &    --   \\
   & DSTFlow~\cite{ren2017unsupervised}           & 10.43 & 3.29 & 12.4 & 4.0  &    --   &   --    & 16.79 & 6.96  &  39\%    &    --   &    --   \\
   & UnFlow-CSS~\cite{Meister:2018:UUL}           & 3.29  & 1.26 & --   & --   &    --   &   --    & 8.10  &  --   & 23.30\%  &    --   &    --   \\
   & OccAwareFlow~\cite{wang2018occlusion}        & 3.55  & --   & 4.2  & --   &    --   &   --    & 8.88  &  --   & 31.2\%   &    --   &    --   \\
   & Back2FutureFlow-None~\cite{Janai2018ECCV}\texttt{*}& --    & --   & --   & --   &    --   &   --    & 6.65  & 3.24  &  --      &    --   &    --   \\
   & Back2FutureFlow-Soft~\cite{Janai2018ECCV}\texttt{*}& --    & --   & --   & --   &    --   &   --    & 6.59  & 3.22  & 22.94\%  & 24.27\% &  22.67\%\\
   & EpipolarFlow~\cite{zhong2019epiflow}         &(2.51) &(0.99)& 3.4  & 1.3  &    --   &   --    &(5.55) &(2.46) & 16.95\%  &    --   &    --   \\
   & Lai~\etal~\cite{lai2019bridging}\texttt{(+Stereo)}& 2.56  & 1.39 & --   & --   &    --   &   --    & 7.13  & 4.31  &  --      &    --   &    --   \\
   & UnOS~\cite{wang2019unos}\texttt{(+Stereo)}   & 1.64  & 1.04 & 1.8  & --   &    --   &   --    & 5.58  &  --   & 18.00\%  &    --   &    --   \\
   \cline{2-13}
   & DDFlow~\cite{Liu:2019:DDFlow}                & 2.35  & 1.02 & 3.0  & 1.1  &  8.86\% &  4.57\% & 5.72  & 2.73  & 14.29\%  & 20.40\% & 13.08\% \\
   & SelFlow~\cite{Liu:2019:SelFlow}\texttt{*}    & 1.69  & 0.91 & 2.2  & 1.0  &  7.68\% &  4.31\% & 4.84  & 2.40  & 14.19\%  & 21.74\% & 12.68\% \\
   & Flow2Stereo~\cite{Liu:2020:Flow2Stereo}\texttt{(+Stereo)}&1.45&\textbf{0.82}&1.7& \textbf{0.9}&7.63\%&4.02\%&3.54&2.12&11.10\%&\textbf{16.67\%}&9.99\% \\
   & DistillFlow \texttt{(trained on Sintel)}     & 2.33  & 1.08 & --   & --   &    --   &  --     & 8.16  & 4.20  &  --      &    --   &    --   \\
   & DistillFlow                                  &\textbf{1.38} & 0.83 &\textbf{1.6}& \textbf{0.9}  & \textbf{7.18\%} &  \textbf{3.91\%} & \textbf{2.93}  &\textbf{1.96}  & \textbf{10.54\%}  & 16.98\% &  \textbf{9.26\%} \\

   \midrule
   \multirow{21}{*}{\rotatebox[origin=c]{90}{Supervised}}
   & FlowNetS~\cite{dosovitskiy2015flownet}       & 7.52  & --   & 9.1  & --   & 44.49\% &   --    &  --   &  --   &  --      &    --   &    --   \\
   & SpyNet~\cite{ranjan2017optical}              & 3.36  & --   & 4.1  & 2.0  & 20.97\% & 12.31\% &  --   &  --   & 35.07\%  & 43.62\% & 33.36\% \\
   & FlowFieldsCNN~\cite{bailer2017cnn}           & --    & --   & 3.0  & 1.2  & 13.01\% & 4.89\%  &  --   &  --   & 18.68\%  & 20.42\% & 18.33\% \\
   & DCFlow~\cite{XRK2017}                        & --    & --   & --   & --   &   --    & --      &  --   &  --   & 14.86\%  & 23.70\% & 13.10\% \\
   & FlowNet2~\cite{ilg2017flownet}               & (1.28)& --   & 1.8  & 1.0  &  8.80\% & 4.82\%  & (2.3) &  --   & 10.41\%  &  8.75\% & 10.75\% \\
   & UnFlow-CSS~\cite{Meister:2018:UUL}           & (1.14)&(0.66)& 1.7  & 0.9  &  8.42\% & 4.28\%  & (1.86)&  --   & 11.11\%  & 15.93\% & 10.15\% \\
   & LiteFlowNet~\cite{hui18liteflownet}          & (1.05)& --   & 1.6  & 0.8  &  7.27\% & 3.27\%  & (1.62)&  --   &  9.38\%  &  7.99\% &  9.66\% \\
   & LiteFlowNet2~\cite{hui2020lightweight2}      & (0.95)& --   & 1.4  & 0.7  &  6.16\% & 2.63\%  & (1.33)&  --   &  7.62\%  &  7.64\% &  7.62\% \\
   & PWC-Net~\cite{sun2018pwc}                    & (1.45)& --   & 1.7  & 0.9  &  8.10\% & 4.22\%  & (2.16)&  --   &  9.60\%  &  9.31\% &  9.66\% \\
   & PWC-Net+~\cite{sun2019models}                & (1.08)& --   & 1.4  & 0.8  &  6.72\% & 3.36\%  & (1.45)&  --   &  7.72\%  &  7.88\% &  7.69\% \\
   & ContinualFlow~\cite{neoral2018continual}     & --    & --   & --   & --   &   --    & --      &  --   &  --   & 10.03\%  & 17.48\% &  8.54\% \\
   & HD$^3$Flow~\cite{yin2019hierarchical}        & (0.81)& --   & 1.4  & 0.7  &  5.41\% & 2.26\%  & (1.31)&  --   &  6.55\%  &  9.02\% &  6.05\% \\
   & IRR-PWC~\cite{Hur:2019:IRR}                  & --    & --   & 1.6  & 0.9  &  6.70\% & 3.21\%  & (1.45)&  --   &  7.65\%  &  7.52\% &  7.68\% \\
   & MFF~\cite{ren2018fusion}\texttt{*}           & --    & --   & 1.7  & 0.9  &  7.87\% & 4.19\%  &  --   &  --   &  7.17\%  &  \textbf{7.25\%} &  7.15\% \\
   & VCN~\cite{yang2019volumetric}                & --    & --   & --   & --   &  --     & --      & (1.16)&  --   &  6.30\%  &  8.66\% &  5.83\% \\
   & SENSE~\cite{jiang2019sense}                  & (1.18)& --   & 1.5  & --   &  --     & 3.03\%  & (2.05)&  --   &  8.16\%  &    --   &    --   \\
   & ScopeFlow~\cite{barhaim2020scopeflow}        & --    & --   & 1.3  & 0.7  &  5.66\% & 2.68\%  &  --   &  --   &  6.82\%  &  7.36\% &  6.72\% \\
   & MaskFlowNet-S~\cite{zhao2020maskflownet}     & --    & --   & \textbf{1.1}  & \textbf{0.6}  &  5.24\% & 2.29\%  &  --   &  --   &  6.81\%  &  8.21\% &  6.53\% \\
   & MaskFlowNet~\cite{zhao2020maskflownet}       & --    & --   & \textbf{1.1}  & \textbf{0.6}  &  \textbf{4.82\%} & \textbf{2.07\%}  &  --   &  --   &  6.11\%  &  7.70\% &  5.79\% \\
   \cline{2-13}

   & SelFlow~\cite{Liu:2019:SelFlow}\texttt{*}    & \textbf{(0.76)}&(0.47)& 1.5  & 0.9  &  6.19\% & 3.32\%  & (1.18)& (0.82)&  8.42\%  &  7.61\% & 12.48\% \\
   & DistillFlow                                  & (0.79)&\textbf{(0.45)}& 1.2  & \textbf{0.6}  &  5.23\% & 2.33\%  & \textbf{(1.14)} & \textbf{(0.74)}&  \textbf{5.94\%}  &  7.96\% &  \textbf{5.53\%} \\
 \bottomrule \end{tabular} }
\vspace{-1ex}
\end{table*}

\begin{table}[ht]
\caption{Quantitative evaluation of optical flow estimation on Sintel dataset. 
}
\label{table:main_sintel_flow}
\centering
\resizebox{0.49\textwidth}{!}{
\begin{tabular}{clcccc}
   \toprule
   &   \multirow{2}{*}{Method} & \multicolumn{2}{c}{Sintel Clean} & \multicolumn{2}{c}{Sintel Final}       \\
                                 \cmidrule(l{3mm}r{3mm}){3-4}        \cmidrule(l{3mm}r{3mm}){5-6} 
   &                                                        & EPE-train & EPE-test & EPE-train & EPE-test  \\
   \midrule
   \multirow{10}{*}{\rotatebox[origin=c]{90}{Unsupervised}}
   & DSTFlow~\cite{ren2017unsupervised}                     &   (6.16)  &   10.41  &   (6.81)  &   11.27   \\
   & UnFlow-CSS~\cite{Meister:2018:UUL}                     &    --     &    --    &   (7.91)  &   10.22   \\
   & OccAwareFlow~\cite{wang2018occlusion}                  &   (4.03)  &    7.95  &   (5.95)  &    9.15   \\
   & Back2FutureFlow-None~\cite{Janai2018ECCV}\texttt{*}    &   (6.05)  &    --    &   (7.09)  &    --     \\
   & Back2FutureFlow-Soft~\cite{Janai2018ECCV}\texttt{*}    &   (3.89)  &    7.23  &   (5.52)  &    8.81   \\
   & EpipolarFlow~\cite{zhong2019epiflow}                   &   (3.54)  &    7.00  &   (4.99)  &    8.51   \\
   \cline{2-6}
   & DDFlow~\cite{Liu:2019:DDFlow}                          &   (2.92)  &    6.18  &   (3.98)  &    7.40   \\
   & SelFlow~\cite{Liu:2019:SelFlow}\texttt{*}              &   (2.88)  &    6.56  &   (3.87)  &    6.57   \\
   & DistillFlow \texttt{(trained on KITTI)}                &    4.21   &    --    &    5.06   &    --     \\                         
   & DistillFlow                                            &\textbf{(2.61)}&\textbf{4.23}&\textbf{(3.70)}&\textbf{5.81}\\

   \midrule
   \multirow{20}{*}{\rotatebox[origin=c]{90}{Supervised}}
   & FlowNetS~\cite{dosovitskiy2015flownet}                 &   (3.66)  &    6.96  &   (4.44)  &    7.76   \\
   & FlowNetC~\cite{dosovitskiy2015flownet}                 &   (3.78)  &    6.85  &   (5.28)  &    8.51   \\
   & SpyNet~\cite{ranjan2017optical}                        &   (3.17)  &    6.64  &   (4.32)  &    8.36   \\
   & FlowFieldsCNN~\cite{bailer2017cnn}                     &    --     &    3.78  &    --     &    5.36   \\
   & DCFlow~\cite{XRK2017}                                  &    --     &    3.54  &    --     &    5.12   \\
   & FlowNet2~\cite{ilg2017flownet}                         &   (1.45)  &    4.16  &   (2.01)  &    5.74   \\
   & LiteFlowNet~\cite{hui18liteflownet}                    &\textbf{(1.35)}&    4.54  &   (1.78)  &    5.38   \\
   & LiteFlowNet2~\cite{hui2020lightweight2}                &   (1.41)  &    3.48  &   (1.83)  &    4.69   \\
   & PWC-Net~\cite{sun2018pwc}                              &   (2.02)  &    4.39  &   (2.08)  &    5.04   \\
   & PWC-Net+~\cite{sun2019models}                          &   (1.71)  &    3.45  &   (2.34)  &    4.60   \\
   & ContinualFlow~\cite{neoral2018continual}               &    --     &    3.34  &    --     &    4.52   \\
   & HD$^3$Flow~\cite{yin2019hierarchical}                  &   (1.70)  &    4.79  &\textbf{(1.17)}&    4.67   \\
   & IRR-PWC~\cite{Hur:2019:IRR}                            &   (1.92)  &    3.84  &   (2.51)  &    4.58   \\
   & MFF~\cite{ren2018fusion}\texttt{*}                     &    --     &    3.42  &    --     &    4.57   \\
   & VCN~\cite{yang2019volumetric}                          &   (1.66)  &    2.81  &   (2.24)  &    4.40   \\
   & SENSE~\cite{jiang2019sense}                            &   (1.54)  &    3.60  &   (2.05)  &    4.86   \\
   & ScopeFlow~\cite{barhaim2020scopeflow}                  &    --     &    3.59  &    --     &\textbf{4.10}   \\
   & MaskFlowNet-S~\cite{zhao2020maskflownet}               &    --     &    2.77  &    --     &    4.38   \\
   & MaskFlowNet~\cite{zhao2020maskflownet}                 &    --     &\textbf{2.52} &    --     &    4.17   \\
   \cline{2-6}
   & SelFlow~\cite{Liu:2019:SelFlow}\texttt{*}              &   (1.68)  &    3.74  &   (1.77)  &    4.26   \\
   & DistillFlow                                            &   (1.63)  &    3.49  &   (1.76)  &\textbf{4.10}\\
 \bottomrule \end{tabular} }
\end{table}

\section{Experiment}
We evaluate and compare our method with state-of-the-art unsupervised and supervised learning methods on standard optical flow benchmarks, including KITTI 2012~\cite{geiger2012we}, KITTI 2015~\cite{menze2015object} and MPI Sintel~\cite{butler2012naturalistic}.

\subsection{Implementation Details}

\mypara{Datasets.}
KITTI 2012 and KITTI 2015 datasets consist of real-world road scenes with sparse ground truth flow.
We use the combination of their multi-view extensions raw datasets for unsupervised training, similar to~\cite{ren2017unsupervised,wang2018occlusion}. To avoid the mixture of training and testing data, we exclude the image pairs with ground truth flow and their neighboring frames (frame number 9-12) as in~\cite{ren2017unsupervised,wang2018occlusion}. For supervised fine-tuning, we use the combination of their official training pairs.

MPI Sintel is a challenging optical flow dataset that contains naturalistic video sequences. It includes a `clean' version and a `final' version, where the "final" version is more realistic and challenging. We extract images from the Sintel movie and manually split them into 145 scenes as the raw dataset, resulting in 10,990 image pairs. For unsupervised training, we first train our model on the raw dataset, then fine-tune on the official Sintel training dataset (including both `clean' and `final' versions). For supervised fine-tuning, we use the combination of `clean' and `final' training pairs with dense annotations.

\mypara{Data preprocessing.}
We rescale the pixel value from [0, 255] to [0, 1] for unsupervised training, while normalizing each channel to be the standard normal distribution for supervised fine-tuning. This is because normalizing the image as input is more robust for illumination changing, which is especially helpful for optical flow estimation. For unsupervised training,  we apply Census Transform~\cite{zabih1994non} to images when computing photometric loss, which has been proved robust for optical flow estimation~\cite{hafner2013census,Meister:2018:UUL}.

We employ similar data augmentation strategies with previous works~\cite{dosovitskiy2015flownet,sun2018pwc,hui18liteflownet,zhao2020maskflownet}, including geometric augmentations (\eg random cropping, scaling, flipping, rotation, translation) and color augmentations (\eg random contrast, brightness, hue, saturation, gamma) for both unsupervised and supervised training. We decrease the degree of augmentations on KITTI datasets. During training, we crop 320 $\times$ 896 patches for KITTI and 384 $\times$ 768 patches for Sintel in all experiments. During testing, we resize the images to 384 $\times$ 1280 for KITTI and 448 $\times$ 1024 for Sintel.

\mypara{Training procedures.}
As shown in Figure~\ref{figure:teaser}, we train multiple teacher models and ensemble their predictions as annotations to obtain more reliable predictions, which will be employed to guide the learning of the student model. However, training too many models will cost a lot of computational resources. In our experiments, we make a compromise and only train two teacher models independently. For each teacher model, we use the last five checkpoints to obtain five flow predictions. As a result, our flow annotations from teacher models are the average of ten predictions.

We train our model with Adam optimizer and set the batch size to be 4 for all experiments. To avoid the trivial solution that all pixels are regarded occluded, we pre-train teacher models for 200k iterations before unsupervised training, where photometric loss is applied to all pixels (including both non-occluded pixels and occluded pixels).

For unsupervised training, we set the initial learning rate as $10^{-4}$ and disrupt the learning rate as suggested by \cite{sun2019models} for a better minimum. In stage 1, we train teacher models with $L_1$, the combination of photometric loss with occlusion handling $L_{pho}$ and edge-aware smoothness loss $L_{smo}$ for 600k iterations. Then we generate the ensembled flow predictions and confidence maps using teacher models and regard them as annotations (just like KITTI with sparse annotations). In stage 2, we initialize our student model with one of the pre-trained teacher models, and then train the student model using the knowledge distillation variant 2 (from the confidence view) for another 600k iterations. Thanks to the simplicity of our loss functions, there is no need to tune hyper-parameters.

For KITTI datasets, the unsupervised training is only performed on the raw datasets and the pre-trained student model serves as initialization for supervised fine-tuning. For Sintel datasets, we conduct unsupervised training on both the raw dataset and the official training set (as shown in Table~\ref{table:main_ablation_sintel}), where the former serves as initialization for supervised fine-tuning and the latter is used for a fair comparison with previous unsupervised methods.

We extend our knowledge distillation idea from unsupervised learning to semi-supervised learning, therefore our supervised fine-tuning also contains two stages. In the first stage, we train our student model on the official training image pairs for 1,000k iterations, where we disrupt the learning rate in a similar way as \cite{sun2019models}. Then we generate flow predictions and confidence maps as self-annotations for raw data. In stage 2, we train our model with the combination of official training pairs with ground truth and raw data with self-annotations for 600k iterations.

\mypara{Evaluation Metrics.}
We consider two widely-used metrics to evaluate optical flow estimation: average endpoint error (EPE, lower is better), percentage of erroneous pixels (Fl, lower is better). Fl is the ranking metric on KITTI benchmarks, and EPE is the ranking metric on Sintel benchmarks. We also report the results over non-occluded pixels (noc)  and occluded pixels (occ) respectively. Stereo matching is the byproduct of our flow models, we use EPE and D1 (share the same definition as Fl) as evaluation metrics. The harmonic average of the precision and recall (F-measure) is used to evaluate the accuracy of occlusion estimation.

\subsection{Main Results}
To alleviate the variance of flow prediction from one single model, we average the results of 10 models (2 independent runs with the last 5 checkpoints for each run). We first run each model to obtain results of different metrics, then average the metric results, which makes our results more reliable especially for ablation studies to analyze the effect of different components. For submission to the benchmark, we average the flow predictions from 2 independent runs.

\mypara{KITTI.} As shown in Table~\ref{table:main_kitti_flow}, DistillFlow achieves the best unsupervised results on both KITTI 2012 and KITTI 2015 datasets and outperforms them by a large margin. Specifically, on the KITTI 2012 training set, DistillFlow achieves EPE-all = 1.38 pixels, outperforming previous best unsupervised monocular method EpipolarFlow~\cite{zhong2019epiflow} by 45\%. Note that EpipolarFlow~\cite{zhong2019epiflow} fine-tunes its model on the KITTI 2012 training set, while DistillFlow is only trained on the raw dataset. DistillFlow outperforms UnOS~\cite{wang2019unos} by 16\%, which utilizes stereo videos and additional constraints during training. On KTTI 2012 benchmark, DistillFlow achieves EPE-all = 1.6 pixels and Fl-all = 7.18\%, not only outperforms all previous unsupervised methods, but also outperforms some famous fully supervised methods, including FlowNet2~\cite{ilg2017flownet}, LiteFlowNet~\cite{hui18liteflownet}, PWC-Net~\cite{sun2018pwc} and MFF~\cite{ren2018fusion}. On the KITTI 2015 dataset, the improvement is also significant. DistillFlow achieves EPE-all = 2.93 pixels on the training set, outperforming Back2FutureFlow~\cite{Janai2018ECCV}(utilizes multiple frames during the training) by 56\%, EpipolarFlow~\cite{zhong2019epiflow} and UnOS~\cite{wang2019unos} by 42\%. On the benchmark, DistillFlow achieves Fl-all = 10.54\%, which is a relative 38\% improvement compared with previous best EpipolarFlow~\cite{zhong2019epiflow} (Fl-all = 16.95\%).

After fine-tuning, DistillFlow also achieves state-of-the-art supervised learning performance on KITTI datasets. Specifically, on KITTI 2012, DistillFlow outperforms PWC-Net+~\cite{sun2019models}, LiteFlowNet2~\cite{hui2020lightweight2}, HD$^3$Flow~\cite{yin2019hierarchical}, IRR-PWC~\cite{Hur:2019:IRR} and ScopeFlow~\cite{barhaim2020scopeflow}, is only inferior to MaskFlowNet~\cite{zhao2020maskflownet}, which incorporates an asymmetric feature matching module and direct occlusion reasoning. On KITTI 2015, DistillFlow achieves Fl-all = 5.94\%, ranking 4th on the benchmark (the top three methods all use stereo data), outperforming all monocular optical flow methods (including the most recent MaskFlowNet~\cite{zhao2020maskflownet} and ScopeFlow~\cite{barhaim2020scopeflow}). This is a remarkable result, since we do not require pre-training our model on any labeled synthetic dataset, while all other state-of-the-art supervised learning methods rely on pre-training on synthetic datasets and follow the specific training schedule (FlyingChairs~\cite{dosovitskiy2015flownet}$\to$ FlyingThings3D~\cite{mayer2016large}).

DistillFlow consistently outperforms our previous work DDFlow~\cite{Liu:2019:DDFlow}, SelFlow~\cite{Liu:2019:SelFlow} and Flow2Stereo~\cite{Liu:2020:Flow2Stereo} (uses stereo data). Specifically, for unsupervised setting, DistillFlow outperforms SelFlow~\cite{Liu:2019:SelFlow} 23\% on KITTI 2012 benchmark and 26\% on KITTI 2015 benchmark; for supervised setting, DistillFlow outperforms SelFlow~\cite{Liu:2019:SelFlow} 20\% on KITTI 2012 and 29\% on KITTI 2015. The improvements mostly come from edge-aware smoothness regularizer and model distillation.

\mypara{MPI Sintel.} Table~\ref{table:main_sintel_flow} summarizes the comparison of DistillFlow with existing unsupervised and supervised learning methods on Sintel. DistillFlow outperforms all previous unsupervised methods for all metrics. On the Sintel Clean benchmark, DistillFlow achieves EPE-all = 4.23 pixels, while previous best method EpipolarFlow~\cite{zhong2019epiflow} achieves EPE-all = 7.00 pixels, around 40\% relative improvement. On the Sintel Final benchmark, DistillFlow achieves 21\% relative improvement. Our initial data distillation method DDFlow~\cite{Liu:2019:DDFlow} even outperforms all other unsupervised methods (including works that come out later than it, \eg EpipolarFlow), demonstrating the effectiveness of our proposed knowledge distillation framework. 
DistillFlow significantly reduces the gap between state-of-the-art supervised learning methods and unsupervised methods.

After supervised fine-tuning, DistillFlow achieves EPE-all = 4.095 pixels on Sintel Final, and outperforms all published method on the benchmark, including our previous winner entry SelFlow~\cite{Liu:2019:SelFlow} and most recent publications \eg ScopeFlow~\cite{barhaim2020scopeflow} and MaskFlowNet~\cite{zhao2020maskflownet}.

Similarly to KITTI, DistillFlow also achieves improvements over our previous method DDFlow~\cite{Liu:2019:DDFlow} and SelFlow~\cite{Liu:2019:SelFlow}, with 32\% relative improvement on Sintel Clean and 12\% relative improvement on Sintel Final.

\mypara{Qualitative results.}
Figure~\ref{figure:sample_unsupervise} shows sample unsupervised results from KITTI and Sintel datasets. DistillFlow can estimate accurate flow and occlusion maps in a totally unsupervised manner. Figure~\ref{figure:benchmark_kitti_unsupervise} and Figure~\ref{figure:benchmark_sintel_unsupervise} show the qualitative comparison with state-of-the-art supervised learning methods on KITTI 2015 and Sintel Final benchmarks respectively. DistillFlow achieves better flow prediction, especially for occluded pixels.

\mypara{Occlusion estimation.}
Following previous works~\cite{wang2018occlusion,Janai2018ECCV}, we evaluate our occlusion estimation performance on both KITTI and Sintel datasets. Note KITTI datasets only have sparse occlusion maps. As shown in  Table~\ref{table:occlusion_estimation}, DistillFlow achieves best occlusion estimation performance on Sintel Clean and Sintel Final datasets over all competing methods. On the KITTI datasets, the ground truth occlusion masks only contain pixels moving out of the image boundary. However, our method will also estimate the occlusions within the image range. Under such settings, our method can achieve comparable performance. DistillFlow consistently outperforms our previous method~\cite{Liu:2019:DDFlow,Liu:2019:SelFlow}, suggesting better occlusion reasoning ability. This also explains why DistillFlow achieves performance improvement.

\begin{table}[t]
\caption{Comparison of occlusion estimation with F-measure.  Note that on KITTI datasets, occlusion only contains pixels moving out of the image boundary and occlusion maps are sparse.}
\label{table:occlusion_estimation}
\centering
\resizebox{0.4\textwidth}{!}{
\begin{tabular}{ l c c c c }
 \toprule   \multirow{2}{*}{Method}                      &   KITTI  &  KITTI &  Sintel  &  Sintel  \\
                                                         &   2012   &  2015  &  Clean   &  Final   \\
	\midrule                                                                 
	MODOF~\cite{xu2011motion}                            &    --    &   --   &   --     &   0.48   \\
	OccAwareFlow~\cite{wang2018occlusion}                &   0.95   &  0.88  &  (0.54)  &  (0.48)  \\
	Back2Future~\cite{Janai2018ECCV}\texttt{*}           &    --    &\textbf{0.91}&  (0.49)  &  (0.44)  \\
	\midrule
	DDFlow~\cite{Liu:2019:DDFlow}                        &   0.94   &  0.86  &\textbf{(0.59)}&  (0.52)  \\
	SelFlow~\cite{Liu:2019:SelFlow}\texttt{*}            &   0.95   &  0.88  &\textbf{(0.59)}&  (0.52)  \\
	DistillFlow                                          &\textbf{0.96}&0.89&\textbf{(0.59)}&\textbf{(0.53)}\\
	\bottomrule \end{tabular}}
\vspace{-1ex}
\end{table}

\subsection{Generalization}
We demonstrate the generalization capability of DistillFlow in three aspects: framework generalization, correspondence generalization and cross-dataset generalization.

\begin{table*}[ht]
\caption{Ablation study for the generalization capability of our proposed distillation framework to FlowNetS and FlowNetC on KITTI and Sintel datasets. Default experimental settings: census transform (yes), occlusion handling (yes), edge-aware smoothness loss (yes).}
\label{table:ablation_generalization_flownet}
\vspace{-1ex}
\centering
\resizebox{\textwidth}{!}{
\begin{tabular}{c c c c c c c c c c c c c c}
  \toprule
     Network  &  Knowledge  & \multicolumn{3}{c}{KITTI 2012} & \multicolumn{3}{c}{KITTI 2015}  &\multicolumn{3}{c}{Sintel Clean}&\multicolumn{3}{c}{Sintel Final}\\
                            \cmidrule(l{3mm}r{3mm}){3-5}     \cmidrule(l{3mm}r{3mm}){6-8}     \cmidrule(l{3mm}r{3mm}){9-11}     \cmidrule(l{3mm}r{3mm}){12-14}
    Backbone  & Distillation    & EPE-all & EPE-noc & EPE-occ & EPE-all  & EPE-noc  & EPE-occ & EPE-all & EPE-noc & EPE-occ  & EPE-all & EPE-noc & EPE-occ \\
  \midrule 
  \multirow{2}{*}{FlowNetS} 
              & \xmark  & 4.26 & 1.53 & 22.34 & 8.85 & 3.82 & 40.63 & (5.05) & (3.09) & (30.01) & (5.38) & (3.38) & (31.00) \\
              & \cmark  &\textbf{2.70}&\textbf{1.38}&\textbf{11.44}&\textbf{6.33}&\textbf{3.44}&\textbf{24.59}&\textbf{(4.20)}&\textbf{(2.36)}&\textbf{(27.66)}& \textbf{(4.83)}&\textbf{(2.90)}&\textbf{(29.49)} \\
  \midrule
  \multirow{2}{*}{FlowNetC} 
              & \xmark  & 3.63 & 1.26 & 19.31 & 8.11 & 3.45 & 37.61 & (4.20) & (2.36) & (27.66) & (4.83) & (2.90) & (29.49) \\ 
              & \cmark  &\textbf{2.18}&\textbf{1.16}&\textbf{8.97}&\textbf{5.47}&\textbf{2.95}&\textbf{21.38}&\textbf{(3.45)}&\textbf{(1.90)}&\textbf{(23.27)}& \textbf{(4.17)}&\textbf{(2.52)}&\textbf{(25.36)} \\ 
  \bottomrule
\end{tabular}
}
\end{table*}
\begin{table*}[ht]
\caption{Ablation study for the generalization capability of our proposed distillation framework to semi-supervised learning on KITTI and Sintel datasets. In this experiment, we split KITTI and Sintel dataset into \textbf{training} and \textbf{validation} datasets and evaluate the performance the \textbf{validation} part.}
\label{table:ablation_distillation_semisupervise}
\vspace{-1ex}
\centering
\resizebox{\textwidth}{!}{
\begin{tabular}{c c c c c c c c c c c c c c c c}
  \toprule
  Semi-Supervised  & \multicolumn{4}{c}{KITTI 2012} & \multicolumn{4}{c}{KITTI 2015}  &\multicolumn{3}{c}{Sintel Clean}&\multicolumn{3}{c}{Sintel Final}  \\
                    \cmidrule(l{3mm}r{3mm}){2-5}     \cmidrule(l{3mm}r{3mm}){6-9}     \cmidrule(l{3mm}r{3mm}){10-12}     \cmidrule(l{3mm}r{3mm}){13-15}
   Learning    & EPE-all & EPE-noc & Fl-all & Fl-noc & EPE-all & EPE-noc & Fl-all  & Fl-noc  & EPE-all & EPE-noc & EPE-occ  & EPE-all & EPE-noc & EPE-occ \\
  \midrule 
  \xmark       &   1.01  &  0.58   & 3.46\% & 1.69\% &  1.50   &\textbf{0.94}&  5.17\% & 3.54\%  &   1.64  &   0.77  &  15.65   &  2.44   &  1.51   & 17.59   \\
  \cmark       &\textbf{0.95}&\textbf{0.57}&\textbf{3.35\%}&\textbf{1.65\%}&\textbf{1.44}&\textbf{0.94}&\textbf{4.96\%}&\textbf{3.51\%}&\textbf{1.56}&\textbf{0.72}&\textbf{15.21}&\textbf{2.38}&\textbf{1.47}&\textbf{17.16}   \\
  \bottomrule
\end{tabular}
}
\end{table*}

\mypara{Framework generalization.}
Our proposed knowledge distillation based self-supervised learning framework is effective for different network structures and is applicable to both unsupervised setting and supervised setting. To verify the former one, apart from PWC-Net based network backbones (as shown in Table~\ref{table:main_ablation_kitti} and Table~\ref{table:main_ablation_sintel}), we also apply our self-supervised learning framework to FlowNetS and FlowNetC (Table~\ref{table:ablation_generalization_flownet}). With knowledge distillation, we achieve more than 30\% relative improvement on average on KITTI datasets for both FlowNestS and FlowNetC, and 15\% relative improvement on Sintel Clean and Final datasets. More importantly, FlowNetS and FlowNetC trained with knowledge distillation achieve EPE-all = 6.33 pixels and 5.47 pixels on KITTI 2015 training dataset, EPE-all = 4.83 pixels and 4.17 pixels on Sintel Final training dataset, which even outperform Back2FutureFlow~\cite{Janai2018ECCV} based on PWC-Net. This also has the same conclusion as \cite{sun2019models}: model matters, so does training. Our knowledge distillation approach enables more effective training. All the above results demonstrate the generalization of our distillation framework to different network structures.

We also extend our knowledge distillation idea from unsupervised learning to supervised fine-tuning, resulting in a semi-supervised learning setting.
The semi-supervised setting enables us to utilize more data. As shown in Table~\ref{table:ablation_distillation_semisupervise}, we achieve improvement on both KITTI and Sintel datasets with knowledge distillation. 

\begin{table*}[t]
\caption{Quantitative evaluation of stereo disparity on KITTI 2012 and KITTI 2015 training datasets (apart from the last columns). 
Our flow model trained on monocular videos achieves comparable performance with state-of-the-art unsupervised stereo learning methods. 
$\star$ denotes that we use their pre-trained model to compute the numbers, while other numbers are from their paper. 
}
\label{table:main_kitti_stereo}
\vspace{-1ex}
\centering
\resizebox{\textwidth}{!}{
\begin{tabular}{ l  c c c c c c c c c c c c }
 \toprule
   \multirow{2}{*}{Method} & \multicolumn{6}{c}{KITTI 2012} & \multicolumn{6}{c}{KITTI 2015} \\
    \cmidrule(l{3mm}r{3mm}){2-7}    \cmidrule(l{3mm}r{3mm}){8-13}
   &  EPE-all & EPE-noc & EPE-occ & D1-all & D1-noc & D1-all (test) & EPE-all & EPE-noc & EPE-occ & D1-all & D1-noc & D1-all (test) \\
   \midrule
   Joung \textit{et al.}~\cite{joung2019unsupervised} & -- & -- & -- & -- & -- & 13.88\% & -- & -- & -- & 13.92\% & -- & --\\
   Godard  \textit{et al.}~\cite{Godard_2017_CVPR} $\star$ & 2.12 & 1.44 & 30.91& 10.41\% & 8.33\% & -- & 1.96 & 1.53 & 24.66 & 10.86\% & 9.22\% & -- \\
   Zhou \textit{et al.}~\cite{Zhou_2017_ICCV}  & -- & -- & -- & -- & -- & -- & -- & -- & -- & 9.41\% & 8.35\% & --\\
   OASM-Net ~\cite{li2018occlusion} & -- & -- & -- & 8.79\% & 6.69\% & 8.60\% & -- & -- & -- & -- & -- & 8.98\%\\
   SeqStereo \textit{et al.}~\cite{yang2018segstereo} $\star$ & 2.37 & 1.63 & 33.62 & 9.64\% & 7.89\% & -- & 1.84 & 1.46 & 26.07 & 8.79\% & 7.7\% & -- \\
   Liu  \textit{et al.}~\cite{liu2019unsupervised} $\star$ & 1.78 & 1.68 & 6.25 & 11.57\% & 10.61\% & -- & 1.52 & 1.48 & 4.23 & 9.57\% & 9.10\% & --\\
   Guo \textit{et al.}~\cite{Guo_2018_ECCV} $\star$ & 1.16 & 1.09 & 4.14 & 6.45\% & 5.82\% & -- & 1.71 & 1.67 & 4.06 & 7.06\% & 6.75\% & --\\
   UnOS~\cite{wang2019unos}  & -- & -- & -- & -- & -- & 5.93\% & -- & -- & -- & \textbf{5.94\%} & -- & 6.67\%\\
   \midrule
   Flow2Stereo~\cite{Liu:2020:Flow2Stereo} & \textbf{1.01}  & \textbf{0.93} & 4.52  & 5.14\% & 4.59\% & 5.11\% & 1.34 & 1.31 & \textbf{2.56} & 6.13\% & \textbf{5.93\%} & \textbf{6.61\%} \\
   DistillFlow (no distillation) & 1.25 & 1.03 & 10.57 & 6.67\% & 4.94\% & -- & 1.44 & 1.30 & 9.13 & 8.19\% & 6.90\% & -- \\
   DistillFlow & 1.02 & 0.95 & \textbf{3.72} & \textbf{4.81\%} & \textbf{4.40\%} & 5.14\% & \textbf{1.23} & \textbf{1.21} & 2.78 & 6.37\% & 6.17\% & 6.81\% \\
\bottomrule \end{tabular} }
\end{table*}

\mypara{Correspondence generalization.}
Stereo disparity, which describes the pixel displacement between two stereo images, can be regarded as a special case of optical flow on the epipolar line. They can both be regarded as a correspondence matching problem. From this point of view, if a model can accurately estimate optical flow, it shall have the ability to accurately estimate stereo disparity as well. Then as a byproduct, we directly use our flow model trained on monocular videos to estimate disparity. Surprisingly, our flow model achieves comparable stereo matching performance with current state-of-the-art unsupervised stereo matching methods. As shown in Table~\ref{table:main_kitti_stereo}, DistillFlow achieves D1-all = 4.81\% on KITTI 2012 training dataset and D1-all = 6.37\% on KITTI 2015 dataset, outperforming some famous stereo matching methods \eg SeqStereo \textit{et al.}~\cite{yang2018segstereo} and Guo \textit{et al.}~\cite{Guo_2018_ECCV}. On the KITTI 2012 and 2015 benchmarks, DistillFlow achieves D1-all 5.14\% and 6.81\%, which are comparable with previous state-of-the-art methods UnOS~\cite{wang2019unos} and our previous method Flow2Stereo~\cite{Liu:2020:Flow2Stereo}. The results on stereo matching demonstrate the generalization of DistillFlow to find correspondences.

\begin{table*}[ht]
\caption{Main ablation study on KITTI training datasets. In this experiment, we employ census transform when computing photometric loss. Note that when data distillation is not employed, model distillation (row 4) denotes ensembling predictions of multiple teacher models.
}
\label{table:main_ablation_kitti}
\vspace{-1ex}
\centering
\resizebox{\textwidth}{!}{
\begin{tabular}{c c c c c c c c c c c c c c}
\toprule
Occlusion & Edge-Aware &    Data      & Model &\multicolumn{5}{c}{KITTI 2012}& \multicolumn{5}{c}{KITTI 2015}\\
                                                     \cmidrule(l{3mm}r{3mm}){5-9}    \cmidrule(l{3mm}r{3mm}){10-14}
Handling  & Smoothness & Distillation & Distillation
                                     & EPE-all & EPE-noc & EPE-occ & Fl-all  & Fl-noc  & EPE-all & EPE-noc & EPE-occ & Fl-all  & Fl-noc  \\

  \midrule
  \xmark & \xmark & \xmark & \xmark &  7.33   &   1.30  &  47.26  & 16.27\% &  5.97\% &   12.49 &   3.59  &  68.82  & 23.07\% & 12.40\% \\
  \cmark & \xmark & \xmark & \xmark &  3.22   &   0.98  &  18.07  & 13.57\% &  4.40\% &    6.57 &   2.88  &  29.87  & 19.90\% & 10.01\% \\
  \cmark & \cmark & \xmark & \xmark &  2.92   &   0.93  &  16.06  & 12.44\% &  3.94\% &    6.45 &   2.59  &  30.90  & 19.08\% &  9.48\% \\
  \cmark & \cmark & \xmark & \cmark &  2.86   &   0.91  &  15.84  & 11.58\% &  3.85\% &    6.36 &   2.52  &  29.76  & 18.24\% &  9.32\% \\
  \cmark & \cmark & \cmark & \xmark &  1.46   &   0.85  &   5.44  &  5.17\% &  3.38\% &    3.20 &   2.08  &  10.28  & 10.05\% &  8.03\% \\
  \cmark & \cmark & \cmark & \cmark &\textbf{1.38}&\textbf{0.83}&\textbf{4.98}&\textbf{4.99\%}&\textbf{3.25\%}&\textbf{2.93}&\textbf{1.96}&\textbf{9.04}&\textbf{9.79\%}&\textbf{7.81\%}\\
  \bottomrule
\end{tabular}
}
\end{table*}
\begin{table*}[ht]
\caption{Main ablation study on Sintel training datasets. In this experiment, we employ census transform when computing photometric loss.  Note that when data distillation is not employed, model distillation (row 4 and row 10) denotes ensembling predictions of multiple teacher models.}
\label{table:main_ablation_sintel}
\vspace{-1ex}
\centering
\resizebox{\textwidth}{!}{
\begin{tabular}{c c c c c c c c c c c c c c c}
  \toprule
  Training & Occlusion & Edge-Aware &  Data        & Model &\multicolumn{5}{c}{Sintel Clean}& \multicolumn{5}{c}{Sintel Final}\\
                                                              \cmidrule(l{3mm}r{3mm}){6-10}    \cmidrule(l{3mm}r{3mm}){11-15}
  Dataset  & Handling  & Smoothness & Distillation & Distillation
                                     & EPE-all & EPE-noc & EPE-occ & Fl-all  & Fl-noc  & EPE-all & EPE-noc & EPE-occ & Fl-all  & Fl-noc \\
  \midrule
  \multirow{5}{*}{Sintel Raw}
  &\xmark & \xmark & \xmark & \xmark &   4.17  &   1.85  &  33.95  & 10.21\% &  5.11\% &   5.36  &   2.86  &  37.30  & 14.46\% & 9.42\% \\
  &\cmark & \xmark & \xmark & \xmark &   3.58  &   1.65  &  28.67  &  8.96\% &  4.17\% &   4.67  &   2.61  &  31.08  & 13.39\% & 8.49\% \\
  &\cmark & \cmark & \xmark & \xmark &   3.29  &   1.54  &  25.70  &  8.25\% &  3.76\% &   4.41  &   2.50  &  28.75  & 12.89\% & 8.18\% \\
  &\cmark & \cmark & \xmark & \cmark &   3.25  &   1.52  &  25.36  &  8.12\% &  3.63\% &   4.36  &   2.46  &  28.22  & 12.45\% & 7.92\% \\
  &\cmark & \cmark & \cmark & \xmark &   3.04  &   1.42  &  23.72  &  7.56\% &  4.58\% &   3.98  &   2.26  &  25.81  & 11.14\% & 6.92\% \\
  &\cmark & \cmark & \cmark & \cmark &\textbf{2.98}&\textbf{1.39}&\textbf{23.43}&\textbf{7.45\%}&\textbf{3.49\%}&\textbf{3.90}&\textbf{2.21}&\textbf{25.52}&\textbf{10.98\%}&\textbf{6.76\%} \\
  \midrule
  \multirow{5}{*}{Sintel Train}
  &\xmark & \xmark & \xmark & \xmark &  (3.93) &  (1.56) & (34.23) & (9.61\%)& (4.45\%)&  (5.19) &  (2.66) & (37.54) &(13.64\%)&(8.54\%)\\
  &\cmark & \xmark & \xmark & \xmark &  (3.22) &  (1.34) & (27.24) & (8.43\%)& (3.51\%)&  (4.40) &  (2.36) & (30.49) &(12.72\%)&(7.44\%)\\
  &\cmark & \cmark & \xmark & \xmark &  (2.93) &  (1.24) & (24.66) & (7.63\%)& (3.15\%)&  (4.17) &  (2.32) & (27.83) &(12.31\%)&(7.62\%)\\
  &\cmark & \cmark & \xmark & \cmark &  (2.89) &  (1.22) & (24.21) & (7.46\%)& (3.08\%)&  (4.12) &  (2.29) & (27.24) &(12.12\%)&(7.51\%)\\
  &\cmark & \cmark & \cmark & \xmark &  (2.66) &  (1.16) & (21.89) & (7.03\%)& (3.14\%)&  (3.76) &  (2.10) & (24.92) &(10.70\%)&(6.55\%)\\
  &\cmark & \cmark & \cmark & \cmark &\textbf{(2.61)}&\textbf{(1.12)}&\textbf{(21.63)}&\textbf{(6.87\%)}&\textbf{(2.99\%)}&\textbf{(3.70)}&\textbf{(2.07)}&\textbf{(24.60)}&\textbf{(10.61\%)}&\textbf{(6.45\%)}\\
  \bottomrule
\end{tabular}
}
\end{table*}

\mypara{Cross-dataset generalization.}
Although deep learning based optical flow methods have outperformed classical methods on challenging benchmarks, their generalization ability is very poor due to limited annotated training data. Therefore, currently learning based methods still cannot apply to many scenes. However, our proposed DistillFlow is a self-supervised learning approach, which can utilize unlimited in-the-wild videos and effectively learn optical flow without requiring any annotations. Since a large collection of unlabeled image sequences can be used, the learned model shall have the strong generalization capability. As shown in Table~\ref{table:main_kitti_flow} (DistillFlow \texttt{(trained on Sintel)}) and Table~\ref{table:main_sintel_flow} (DistillFlow \texttt{(trained on KITTI)}), we use models trained on Sintel to estimate flow on KITTI and vice versa. Surprisingly, for Sintel~$\to$~KITTI, DistillFlow achieves EPE = 2.33 pixels on KITTI 2012 training dataset, outperforming previous state-of-the-art unsupervised learning method Back2FutureFlow~\cite{Janai2018ECCV}. On KITTI 2015, DistillFlow outperforms OccAwareFlow~\cite{wang2018occlusion} is also comparable with Back2FutureFlow~\cite{Janai2018ECCV}. For KITTI~$\to$~Sintel, DistillFlow achieves EPE = 5.06 pixels on Sintel Final training dataset, which outperforms Back2FutureFlow~\cite{Janai2018ECCV} and is comparable with EpipolarFlow~\cite{zhong2019epiflow}. This is indeed a remarkable result, since KITTI datasets only have street views while the Sintel dataset contains many complex scenes. Our model trained only on KITTI datasets achieves even better or comparable performance with state-of-the-art unsupervised learning methods fine-tuned on Sintel dataset. This fully demonstrates the cross-dataset generalization capability of our model. Moreover, since our knowledge distillation method can work well without requiring any labeled data, we can actually train it on a specific scene to achieve better performance. This makes DistillFlow effective for a wider range of applications.

\subsection{Ablation Study}
\begin{figure*}[t]
\centering
\includegraphics[width=1\textwidth]{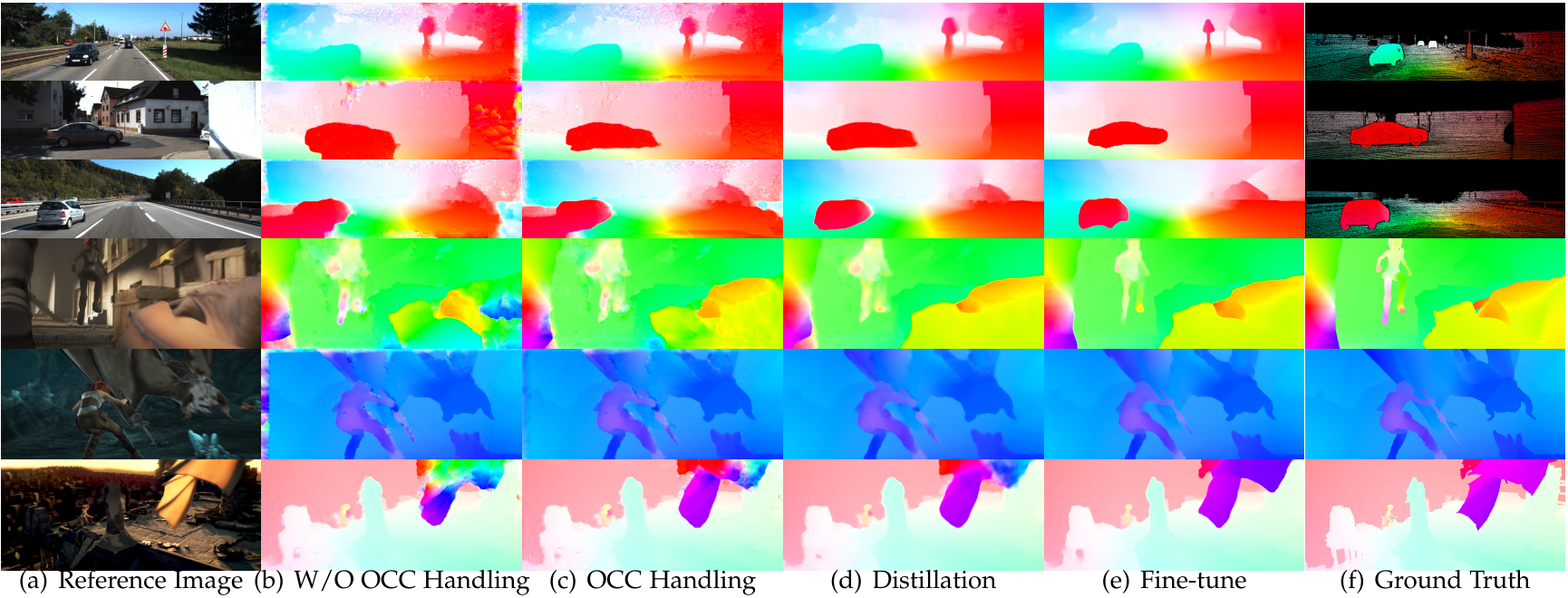}
\vspace{-4ex}
\caption{Ablation study on KITTI 2015 (top 3) and Sintel Final training datasets (bottom 3). (b) and (c) are the results of without and with occlusion handling. (d) shows that results with knowledge distillation and (f) is the supervised fine-tuned results. With knowledge distillation, the flow looks more smooth. After fine-tuning, more details are preserved. Zoom in for details.}
\label{figure:ablation}
\end{figure*}

We conduct a thorough ablation study to demonstrate the effectiveness of different components proposed by DistillFlow. Figure~\ref{figure:ablation} shows visual comparisons.

\mypara{Occlusion handling.}
As shown in Table~\ref{table:main_ablation_kitti} (row 1 vs. row 2) and Table~\ref{table:main_ablation_sintel} (row 1 vs. row 2 and row 7 vs. row 8), occlusion handling can improve the flow estimation performance on all datasets for all metrics. This is because that brightness constancy assumption does not hold for occluded pixels. Occlusion handling can reduce the misleading guidance information provided by occluded pixels, which makes the  model easier to learn good correspondence.

\mypara{Edge-aware smoothness.}
As shown in Table~\ref{table:main_ablation_kitti} (row 2 vs. row 3) and Table~\ref{table:main_ablation_sintel} (row 2 vs. row 3 and row 8 vs. row 9), edge-aware smoothness regularizer can also consistently improve the performance on all datasets. This is because photometric loss is not informative in homogeneous regions and cannot handle occlusions. The spatial smooth assumption regularizes the flow to be locally similar, which helps predict flow of those homogeneous or texture-less regions. Smoothness can be regarded as a regularizer for some occluded pixels, since it makes the prediction of occluded pixels similar to the neighborhood of non-occluded pixels. However, it is just a very weak regularizer, therefore we propose knowledge distillation to more effectively learn optical flow of occluded pixels.

\begin{table*}[ht]
\caption{Ablation study of different knowledge distillation strategies on KITTI and Sintel datasets. For `v1', `v2' and `v3', we use knowledge distillation variant 1 (from occlusion view and split pixels in occluded and non-occluded), while for `v2' we use variant 2 (from confidence view). `v1' denotes distillation used in DDFlow~\cite{Liu:2019:DDFlow}, `v2' denotes distillation used in SelFlow~\cite{Liu:2019:SelFlow}, `v3' and `v4' denotes distillation with more challenging transformations as in Flow2Stereo~\cite{Liu:2020:Flow2Stereo}. Default experimental settings: census transform (yes), occlusion handling (yes), edge-aware smoothness loss (yes).}
\label{table:ablation_distillation}
\vspace{-1ex}
\centering
\resizebox{\textwidth}{!}{
\begin{tabular}{c c c c c c c c c c c c c c c}
  \toprule
    Knowledge      & \multicolumn{3}{c}{KITTI 2012} & \multicolumn{3}{c}{KITTI 2015} &\multicolumn{3}{c}{Sintel Clean}&\multicolumn{3}{c}{Sintel Final}\\
                  \cmidrule(l{3mm}r{3mm}){2-4}     \cmidrule(l{3mm}r{3mm}){5-7}    \cmidrule(l{3mm}r{3mm}){8-10}     \cmidrule(l{3mm}r{3mm}){11-13}
  Distillation & EPE-all & EPE-noc & EPE-occ & EPE-all & EPE-noc & EPE-occ & EPE-all & EPE-noc & EPE-occ  & EPE-all & EPE-noc & EPE-occ \\
  \midrule
  no           &   2.92  &   0.93  &  16.06  &   6.45  &   2.59  &   30.90 &  (2.93) & (1.24)  & (24.66)  &  (4.17) &  (2.32) & (27.83) \\
  v1           &   1.62  &   0.87  &   6.21  &   3.88  &   2.19  &   15.24 &  (2.76) & (1.16)  & (22.98)  &  (3.94) &  (2.16) & (25.72) \\
  v2           &   1.54  &   0.87  &   5.77  &   3.57  &   2.10  &   12.88 &  (2.71) & (1.18)  & (22.51)  &  (3.87) &  (2.19) & (25.38) \\
  v3           &   1.41  &   0.85  &   5.12  &   3.12  &   2.01  &    9.48 &  (2.63) & \textbf{(1.12)}  & (21.72)  &  (3.74) &  (2.09) & (24.81) \\ 
  v4           &\textbf{1.38}&\textbf{0.83}&\textbf{4.98}&\textbf{2.93}&\textbf{1.96}&\textbf{9.04}&\textbf{(2.61)}&\textbf{(1.12)}&\textbf{(21.63)}&\textbf{(3.70)}&\textbf{(2.07)}&\textbf{(24.60)} \\ 
  \bottomrule
\end{tabular}
}
\end{table*}
\begin{table*}[ht]
\caption{Ablation study of photometric losses. Default experiment settings: edge-aware smoothness loss (no), occlusion handling (yes), distillation (no).}
\label{table:ablation_photometric}
\vspace{-1ex}
\centering
\resizebox{\textwidth}{!}{
\begin{tabular}{c c c c c c c c c c c c c c c}
  \toprule
     Pixel   & \multirow{2}{*}{SSIM} &  Census      & \multicolumn{3}{c}{KITTI 2012} & \multicolumn{3}{c}{KITTI 2015} 
                                                    &\multicolumn{3}{c}{Sintel Clean}&\multicolumn{3}{c}{Sintel Final}\\
                                                       \cmidrule(l{3mm}r{3mm}){4-6}     \cmidrule(l{3mm}r{3mm}){7-9}
                                                       \cmidrule(l{3mm}r{3mm}){10-12}     \cmidrule(l{3mm}r{3mm}){13-15}
  Brightness &                       & Transform 
                           & EPE-all & EPE-noc & EPE-occ & EPE-all  & EPE-noc  & EPE-occ & EPE-all & EPE-noc & EPE-occ  & EPE-all & EPE-noc & EPE-occ \\
  \midrule
  \cmark & \xmark & \xmark &  6.58   &  2.03   &  36.69  &   10.63  &   3.99   &  52.62  &  (4.44) &  (2.12) & (34.16)  &  (6.86) &  (4.42) & (38.05) \\
  \cmark & \cmark & \xmark &  5.75   &  1.09   &  36.62  &    9.85  &   3.14   &  52.32  &  (4.15) &  (1.98) & (32.00)  &  (6.22) &  (3.71) & (38.29) \\
  \xmark & \xmark & \cmark &\textbf{3.22}&\textbf{0.98}&\textbf{18.07}&\textbf{6.57}&\textbf{2.88} &\textbf{29.87}&\textbf{(3.22)}&\textbf{(1.34)}&\textbf{(27.24)}&\textbf{(4.40)}&\textbf{(2.36)}&\textbf{(30.49)} \\ 
  \bottomrule
\end{tabular}
}
\end{table*}

\mypara{Data distillation.}
Our proposed knowledge distillation approach contains both data distillation and model distillation. Among them, data distillation is the key point, where we create challenging transformations and let confident predictions to supervise less confident predictions. As shown in Table~\ref{table:main_ablation_kitti} (row 3 vs. row 5), we reduce EPE-all from 2.92 pixels to 1.46 pixels, from 6.45 pixels to 3.20 pixels on KITTI 2012 and KITTI 2015 datasets, with 50\% relative improvement on average. The improvement over occluded pixels is even more significant, with 62\% relative improvement on average. This is because our proposed data distillation enables the model to have the ability to effectively learn optical flow of occluded pixels for the first time.

As shown in Table~\ref{table:main_ablation_sintel} (row 3 vs. row 5 and row 9 vs. row 11), we also achieve great improvement on Sintel datasets. Specifically, we achieve 9\% average relative improvement on both Sintel Clean and Sintel Final. All these results demonstrate the effectiveness of our proposed data distillation approach.

\mypara{Model distillation.}
Since flow predictions from one single teacher model have large variance, we propose model distillation to ensemble flow predictions of multiple teacher models. Model distillation provides more reliable confident flow predictions, which therefore improves the performance. As shown in Table~\ref{table:main_ablation_kitti} (row 3 vs. row 4) and Table~\ref{table:main_ablation_sintel} (row 3 vs. row 4 and row 9 vs. row 10), the average of multiple teacher models has higher accuracy than one teacher model. Consequently, data distillation equipped with model distillation further improves the performance as show in Table~\ref{table:main_ablation_kitti} (row 5 vs. row 6) and Table~\ref{table:main_ablation_sintel} (row 5 vs. row 6 and row 11 vs. row 12).

\mypara{Knowledge distillation strategies.}
In Section~\ref{section:challenging_transformations}, we introduce two variants for knowledge distillation: from occlusion view and from confidence view, resulted in two knowledge distillation strategies `v3' and `v4' in Table~\ref{table:ablation_distillation}. We also provide two knowledge distillation strategies as in our previous work DDFlow~\cite{Liu:2019:DDFlow} and SelFlow~\cite{Liu:2019:SelFlow}, denoted as `v1' and `v2'. Both `v1' and `v2' are from the occlusion view. As shown in Table~\ref{table:ablation_distillation}, all of these four kinds of knowledge distillation strategies can greatly improve the performance, especially for occluded pixels. Comparing `v1' and `v2', we show that superpixel occlusion hallucination can handle occlusion in a wider range of cases. compared with `v2', we add more challenging transformations for `v3', such as geometric transformations and color transformations. As a result, we achieve slight performance improvements. However, most performance gains come from occlusion hallucination techniques. Comparing `v3' and 'v4', we show that it does not make much difference to distinguish occluded or non-occluded pixels during the knowledge distillation stage. This is because forward-backward consistency only predicts confident or less confident flow predictions, but not occluded or non-occluded pixels. 

\mypara{Photometric losses.}
When computing photometric loss, certain transformations are usually applied to the images to make them more robust for illumination changes. As a result, different papers employ different strategies, \eg raw pixel intensity~\cite{wang2018occlusion,Janai2018ECCV}, SSIM~\cite{yin2018geonet} and census transform~\cite{Meister:2018:UUL}. To analyze the effect of different photometric loss, we make a comparison in Table~\ref{table:ablation_photometric}, where SSIM is better than the raw pixel intensity and census transform achieves the best performance. This is because census transform is specially designed to handle the change of illumination. However, we believe census transform is not the optimal transformation for optical flow estimation. Exploring more robust transformations when computing the photometric difference is a potential direction for future research.

\section{Conclusion}
We have presented DistillFlow, a knowledge distillation approach to effectively learning optical flow in a self-supervised manner. To this end, we train multiple teacher models and a student model, where teacher models are used to generate confident flow predictions, which will then be employed to guide the learning of the student model. To make the knowledge distillation effective, we create three types of challenging transformations: occlusion hallucination-based transformations, geometric transformations and color transformations. We show that the key factors for performance gains are generating hallucinated occlusions as well as less confident predictions. With knowledge distillation, DistillFlow achieves the best performance on both KITTI and Sintel datasets and outperforms previous unsupervised methods by a large margin, especially for occluded pixels. More importantly, our self-supervised pre-trained model provides an excellent initialization for supervised fine-tuning.  We show that it is possible to completely remove the reliance of pre-training on synthetic labeled datasets, and achieve superior performance by self-supervised pre-training on unlabeled data. Furthermore, we demonstrate the generalization capability of DistillFlow in three aspects: framework generalization, correspondence generalization and cross-dataset generalization. 
Going forward, our  results suggest that our knowledge distillation technique may be a promising direction for advancing other vision tasks like depth estimation or semantic segmentation.

\ifCLASSOPTIONcompsoc
  \section*{Acknowledgments}
\else
  \section*{Acknowledgment}
\fi
The work described in this paper was partially supported by the National Key Research and Development Program of China (No. 2018AAA0100204) and Research Grants Council of the Hong Kong Special Administrative Region, China (CUHK 14210717 of the General Research Fund).

\bibliographystyle{IEEEtran}
\bibliography{egbib}


\ifCLASSOPTIONcaptionsoff
  \newpage
\fi




\end{document}